\documentclass[review, 3p, times]{elsarticle}
\usepackage{graphics}
\usepackage{latexsym}
\usepackage{amsfonts}
\usepackage{amssymb}
\usepackage{amsthm}
\usepackage{amsmath}
\usepackage{hyperref}
\usepackage{pdfpages}
\usepackage[ruled,vlined]{algorithm2e}
\usepackage{multirow}
\usepackage{color}
\usepackage{tikz}
\usetikzlibrary{shapes.gates.logic.US,trees,positioning,arrows}
\usepackage{graphicx}
\usepackage{floatrow}
\usepackage[position=top]{subfig}
\usepackage{placeins}
\usepackage{enumitem}
\usepackage{framed,multirow}
\usepackage{url}
\usepackage{xcolor}
\definecolor{newcolor}{rgb}{.8,.349,.1}

\begin{document}
\floatsetup[figure]{style=plain,subcapbesideposition=top}
\begin{frontmatter}
\title{SubTSBR to tackle high noise and outliers for data-driven discovery of differential equations}

\author[purdue1]{Sheng Zhang}
\author[purdue1,purdue2]{Guang Lin\corref{cor}}

\address[purdue1]{Department of Mathematics, Purdue University, West Lafayette, IN 47907, USA}
\address[purdue2]{School of Mechanical Engineering, Purdue University, West Lafayette, IN 47907, USA}

\cortext[cor]{Corresponding author.}
\ead{guanglin@purdue.edu}

\begin{abstract}
Data-driven discovery of differential equations has been an emerging research topic. We propose a novel algorithm subsampling-based threshold sparse Bayesian regression (SubTSBR) to tackle high noise and outliers. The subsampling technique is used for improving the accuracy of the Bayesian learning algorithm. It has two parameters: subsampling size and the number of subsamples. When the subsampling size increases with fixed total sample size, the accuracy of our algorithm goes up and then down. When the number of subsamples increases, the accuracy of our algorithm keeps going up. We demonstrate how to use our algorithm step by step and compare our algorithm with threshold sparse Bayesian regression (TSBR) for the discovery of differential equations. We show that our algorithm produces better results. We also discuss the merits of discovering differential equations from data and demonstrate how to discover models with random initial and boundary condition as well as models with bifurcations. The numerical examples are: (1) predator-prey model with noise, (2) shallow water equations with outliers, (3) heat diffusion with random initial and boundary condition, and (4) fish-harvesting problem with bifurcations.
\end{abstract}

\end{frontmatter}

\section{Introduction}
\label{intro}
The search for physical laws has been a fundamental aim of science for centuries. The physical laws are critical to the understanding of natural phenomena and the prediction of future dynamics. They are either derived by other known physical laws or generalized based on empirical observations of physical behavior. We focus on the second task, which is also called data-driven discovery of governing physical laws. It deals with the case where experimental data are given while the governing physical model is unclear. Traditional methods for discovering physical laws from data include interpolation and regression. Suppose $x:\mathbb{R}\to \mathbb{R}$ is an unknown physical law. Given data $\left\{t_i,x(t_i)\right\}_{i=1}^N$, traditional methods approximate the expression of $x(t)$ in terms of a class of functions of $t$. In this paper, we follow a different approach, data-driven discovery of governing differential equations, which adopts the strategy: first, discover the differential equations that $x(t)$ satisfies; second, solve the differential equations analytically or numerically. This discovery pattern is applicable to a larger class of models than traditional methods and derives the governing differential equations, which provide insights to the governing physical laws behind the observations \cite{zhang_robust_2018}. Many fundamental laws are formulated in the form of differential equations, such as Maxwell equations in classical electromagnetism, Einstein field equations in general relativity, Schrodinger equation in quantum mechanics, Navier-Stokes equations in fluid dynamics, Boltzmann equation in thermodynamic, predator-prey equations in biology, and Black-Scholes equation in economics. While automated techniques for generating and collecting data from scientific measurements are more and more precise and powerful, automated processes for extracting knowledge in analytic forms from data are limited \cite{schmidt_distilling_2009}. Our goal is to develop automated algorithms for extracting the governing differential equations from data.

Consider a differential equation of the form
\begin{equation}\label{intro1}
  \frac{dx}{dt} = f(t,x),
\end{equation}
with the unknown function $f(t,x)$. Given the data $\left\{t_i,x_i,x'_i\right\}_{i=1}^N$ collected from a space governed by this differential equation, where $x_i = x(t_i)$ and $x'_i=(dx/dt)(t_i)$, automated algorithms for deriving the expression of $f(t,x)$ are studied from various approaches (in application, if the gradients are not given, they can be approximated numerically). One of the approaches assumes that $f(t,x)$ is a linear combination of simple functions of $t$ and $x$. First, construct a moderately large set of basis-functions that may contain all the terms of $f(t,x)$; then, apply algorithms to select a subset that is exactly all the terms of $f(t,x)$ from the basis-functions and estimate the corresponding weights in the linear combination.

Suppose the basis-functions are chosen as $f_1(t,x),f_2(t,x),\dots,f_M(t,x)$. Then we need to estimate the weights $w_1,w_2,\dots,w_M$ in the following linear combination:
\begin{equation}\label{intro2}
    \frac{dx}{dt} = w_1 f_1(t,x) + w_2 f_2(t,x) + \cdots + w_M f_M(t,x).
\end{equation}
Given the data $\left\{t_i,x_i,x'_i\right\}_{i=1}^N$, where $x_i = x(t_i)$ and $x'_i=(dx/dt)(t_i)$, the above problem becomes a regression problem as follows:
\begin{equation}\label{intro3}
    \left[
  \begin{array}{c}
    x_1' \\
    x_2' \\
    \vdots \\
    x_N'
  \end{array}
\right] = \left[
  \begin{array}{cccc}
    f_1(t_1,x_1) & f_2(t_1,x_1) & \cdots & f_M(t_1,x_1) \\
    f_1(t_2,x_2) & f_2(t_2,x_2) & \cdots & f_M(t_2,x_2) \\
    \vdots & \vdots & \ddots & \vdots \\
    f_1(t_N,x_N) & f_2(t_N,x_N) & \cdots & f_M(t_N,x_N) \\
  \end{array}
\right]
\left[
  \begin{array}{c}
    w_1 \\
    w_2 \\
    \vdots \\
    w_M
  \end{array}
\right] + \epsilon,
\end{equation}
where $\epsilon=\left[\epsilon_1,\epsilon_2,\dots,\epsilon_N\right]^\text{T}$ is the model error. Let
\begin{eqnarray}
  \eta &=& \left[x_1',\dots,x_N'\right]^\text{T} \\
  \Phi &=& \left[
    \begin{array}{ccc}
      f_1(t_1,x_1) & \cdots & f_M(t_1,x_1) \\
      \vdots & \ddots & \vdots \\
      f_1(t_N,x_N) & \cdots & f_M(t_N,x_N) \\
    \end{array}
  \right] \\
  \mathbf{w} &=& \left[w_1,\dots,w_M\right]^\text{T}.
\end{eqnarray}
Equation (\ref{intro3}) may be written in the vector form as follows:
\begin{equation}\label{intro4}
  \eta = \Phi \mathbf{w} + \epsilon.
\end{equation}
Now the problem is to estimate the weight-vector $\mathbf{w}$ given a known vector $\eta$ and a known matrix $\Phi$.

Since many physical systems have few terms in the equations, the set of basis-functions usually has more terms than $f(t,x)$: $M > \#\{\text{terms in }f(t,x)\}$, which suggests the use of sparse methods to select the subset of basis-functions and estimate the weights. These sparse methods include linear algebraic methods, optimization methods, and Bayesian methods. Examples of linear algebraic methods for discovering differential equations are sequential threshold least squares (also called sparse identification of nonlinear dynamics (SINDy)) \cite{brunton_discovering_2016}, which does least-square regression and eliminates the terms with small weights iteratively, and its improvement sequential threshold ridge regression \cite{rudy_data-driven_2017}. An advantage of linear algebraic methods is that they are easy to implement. An example of optimization methods for discovering differential equations is LASSO (least absolute shrinkage and selection operator) \cite{tibshirani_regression_1996, schaeffer_learning_2017, kang_ident_2019}. LASSO solves the following optimization problem:
\begin{equation}\label{intro5}
    \min_{\mathbf{w}}\left\{\frac{1}{2N}||\eta - \Phi \mathbf{w}||_2^2 + \lambda||\mathbf{w}||_1\right\},
\end{equation}
where the regularization parameter $\lambda$ may be fitted by cross-validation. Recently, an improvement of LASSO called SR3 is proposed in \cite{zheng_unified_2019} and might produce better results. An advantage of optimization methods is that they have an explicit objective function to optimize. An example of Bayesian methods for discovering differential equations is threshold sparse Bayesian regression \cite{zhang_robust_2018}, which calculates the posterior distribution of $\mathbf{w}$ given the data and then filters out small weights, iteratively until convergence (Algorithm \ref{algorithm_tsbr}). An advantage of Bayesian methods is that they provide posterior distributions for further analysis. A comparison in \cite{zhang_robust_2018} shows that threshold sparse Bayesian regression is more accurate and robust than sequential threshold least squares and LASSO.

The same mechanism as above also applies to the discovery of general differential equations including higher-order differential equations and implicit differential equations \cite{zhang_robust_2018}, besides the differential equations of the form (\ref{intro1}). Nevertheless, the mechanism is described in the pattern (\ref{intro1}) here for convenience and simplification, so that more attention is given to the essence of the algorithm itself. In addition, to apply the algorithm to real-world problems, dimensional analysis can be incorporated in the construction of the basis-functions \cite{zhang_robust_2018}. Any physically meaningful equation has the same dimensions on every term, which is a property known as dimensional homogeneity. Therefore, when summing up terms in the equations, the addends should be of the same dimension.

Sparse regression methods for data-driven discovery of differential equations are also developed in other papers recently with a wide range of applications, for example, inferring biological networks \cite{mangan_inferring_2016}, sparse identification of a predator-prey system \cite{dam_sparse_2017}, model selection via integral terms \cite{schaeffer_sparse_2017}, extracting high-dimensional dynamics from limited data \cite{schaeffer_extracting_2017}, recovery of chaotic systems from highly corrupted data \cite{tran_exact_2017}, model selection for dynamical systems via information criterion \cite{mangan_model_2017}, model predictive control in the low-data limit \cite{kaiser_sparse_2017}, sparse learning of  stochastic dynamical systems \cite{boninsegna_sparse_2018}, model selection for hybrid dynamical systems \cite{mangan_model_2018}, identification of parametric partial differential equations \cite{rudy_data-driven_2018}, extracting structured dynamical systems with very few samples \cite{schaeffer_extracting_2018}, constrained Galerkin regression \cite{loiseau_constrained_2018}, rapid model recovery \cite{quade_sparse_2018}, convergence of the SINDy algorithm \cite{zhang_convergence_2018}. Moreover, other methods for data-driven discovery of differential equations are proposed as well, for instance, deep neural networks \cite{rudy_deep_2018, raissi_physics_2017-1, raissi_physics_2017} and Gaussian process \cite{raissi_machine_2017}. One of the advantages of the sparse regression methods is the ability to provide explicit formulas of the differential equations, from which further analysis on the systems may be performed, while deep neural networks usually provide ``black boxes'', in which the mechanism of the systems is not very clearly revealed. Another advantage of the sparse regression methods is that they do not require too much prior knowledge of the differential equations, while Gaussian process methods have restrictions on the form of the differential equations and are used to estimate a few parameters.

Previous developments and applications based on sparse regression methods mostly employ either sequential threshold least squares or LASSO, or their variations. One of the reasons why data-driven discovery of differential equations has not yet been applied to industry is the instability of its methods. Previous methods require the data of very high quality, which is usually not the case in industry. Although threshold sparse Bayesian regression is more accurate and robust \cite{zhang_robust_2018}, its performance is still unsatisfactory if the provided data are of high noise or contain outliers. However, it provides a model-selection criterion that allows us to conduct further research and make improvements. In this paper, we develop a subsampling-based technique for improving the threshold sparse Bayesian regression algorithm, so that the new algorithm is robust to high noise and outliers. Note that subsampling methods are usually employed for estimating statistics \cite{efron_jackknife_1981} or speeding up algorithms \cite{rudy_data-driven_2017} in the literatures, but the subsampling method in this paper is used for improving the accuracy of the Bayesian learning algorithm. In practice, denoising techniques can be used to reduce part of the noise and outliers in the data before our algorithm is performed.

The remainder of this paper is structured as follows. In Section \ref{tsbr}, we give a review of the threshold sparse Bayesian regression algorithm. In Section \ref{sub}, we introduce our new subsampling-based algorithm. In Section \ref{example}, we use two examples to detail the mechanism of our algorithm and demonstrate how to apply our algorithm to discover models. In Section \ref{app}, we discuss the merits of discovering differential equations from data and demonstrate how to discover models with random initial and boundary condition as well as models with bifurcations. Finally, the conclusion is given in Section \ref{summary}.

\section{Review of threshold sparse Bayesian regression}
\label{tsbr}
This section is a review of the threshold sparse Bayesian regression algorithm (TSBR) proposed in our previous work \cite{zhang_robust_2018}.

\subsection{Bayesian hierarchical model setup}
\begin{figure}[t]
  \centering
  \begin{tikzpicture}[scale=1, transform shape]
    \begin{scope}[very thick, node distance=2cm, on grid, >=stealth',
		         node/.style={circle,draw,fill=orange!40}]
      \node [node] (eta)  {$\eta$};

      \node [node] (w1)	    [above=of eta, xshift=-2cm]  {$w_1$}       edge [->] (eta);
      \node [node] (w2)	    [above=of eta, xshift=-1cm]  {$w_2$}       edge [->] (eta);
      \node        (dotw)    [above=of eta]               {$\cdots$}    edge [->] (eta);
      \node [node] (wM)	    [above=of eta, xshift=1cm]   {$w_M$}       edge [->] (eta);
      \node [node] (sigma2)  [above=of eta, xshift=2cm]   {$\sigma^2$}  edge [->] (eta);

      \node [node] (a1)	  [above=of w1]    {$\alpha_1$} edge [->] (w1);
      \node [node] (a2)	  [above=of w2]    {$\alpha_2$} edge [->] (w2);
      \node        (dota)  [above=of dotw]  {$\cdots$}   edge [->] (dotw);
      \node [node] (aM)	  [above=of wM]    {$\alpha_M$} edge [->] (wM);
    \end{scope}
  \end{tikzpicture}
  \caption{Graphical structure of the sparse Bayesian model.}\label{tsbr4}
\end{figure}
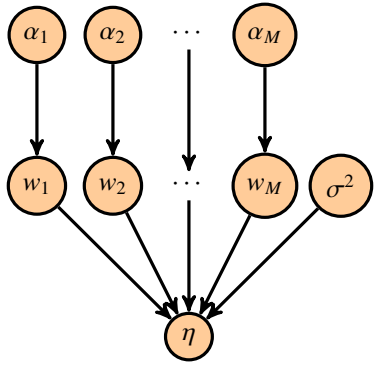

Let $\eta$ be a known $N\times 1$ vector, $\Phi$ be a known $N\times M$ matrix, $\mathbf{w}=\left[w_1, w_2, \dots, w_M\right]^\text{T}$ be the weight-vector to be estimated sparsely, and $\epsilon$ be the model error:
\begin{equation}\label{tsbr1}
  \eta = \Phi \mathbf{w} + \epsilon.
\end{equation}
TSBR adopts a sparse Bayesian framework based on relevance vector machine (RVM) \cite{tipping_sparse_2001}, which is motivated by automatic relevance determination (ARD) \cite{mackay_bayesian_1996, neal_bayesian_1996}, to estimate the weight-vector $\mathbf{w}$. Similar frameworks have applications in compressed sensing \cite{ji_bayesian_2008, ji_multitask_2009, babacan_bayesian_2010}. The Bayesian framework assumes that the model errors are modeled as independent and identically distributed zero-mean Gaussian with variance $\sigma^2$. The variance may be specified beforehand, but here it is fitted by the data. The model gives a multivariate Gaussian likelihood on the vector $\eta$:
\begin{equation}\label{tsbr2}
  p\left(\eta|\mathbf{w},\sigma^2\right) = \left(2\pi \sigma^2\right)^{-N/2}\exp\left\{-\frac{\left||\eta-\Phi\mathbf{w}\right||^2}{2\sigma^2}\right\}.
\end{equation}
Next, a Gaussian prior is introduced over the weight-vector. The prior is governed by a set of hyper-parameters, one hyper-parameter associated with each component of the weight-vector:
\begin{equation}\label{tsbr3}
  p\left(\mathbf{w}|\alpha\right) = \prod_{j=1}^M \mathcal{N}\left(w_j|0,\alpha_j^{-1}\right),
\end{equation}
where $\alpha = \left[\alpha_1,\alpha_2,\dots,\alpha_M\right]^\text{T}$. The values of the hyper-parameters are estimated from the data. See Figure \ref{tsbr4} for the graphical structure of this model.

\subsection{Inference}
The posterior over all unknown parameters given the data can be decomposed as follows:
\begin{equation}\label{tsbr5}
  p\left(\mathbf{w},\alpha,\sigma^2|\eta\right) = p\left(\mathbf{w}|\eta,\alpha,\sigma^2\right)p\left(\alpha,\sigma^2|\eta\right).
\end{equation}
As analytic computations cannot be performed in full, TSBR approximates $p\left(\alpha,\sigma^2|\eta\right)$ using the Dirac delta function at the maximum likelihood estimation:
\begin{eqnarray}\label{tsbr6}
  \left(\hat{\alpha}_\text{ML},\hat{\sigma}^2_\text{ML}\right) &=& \arg\max_{\alpha,\sigma^2} \left\{p\left(\eta|\alpha,\sigma^2\right)\right\}  \nonumber \\
  &=& \arg\max_{\alpha,\sigma^2} \left\{\int p\left(\eta,\mathbf{w}|\alpha,\sigma^2\right)d\mathbf{w}\right\}  \nonumber \\
  &=& \arg\max_{\alpha,\sigma^2} \left\{\int p\left(\eta|\mathbf{w},\sigma^2\right)p\left(\mathbf{w}|\alpha\right)d\mathbf{w}\right\}  \nonumber \\
  &=& \arg\max_{\alpha,\sigma^2} \left\{\left(2\pi\right)^{-N/2}\left|\sigma^2\mathbf{I} + \Phi\mathbf{A}^{-1}\Phi^\text{T}\right|^{-1/2}\exp\left\{-\frac{1}{2}\eta^\text{T}\left(\sigma^2\mathbf{I} + \Phi\mathbf{A}^{-1}\Phi^\text{T}\right)^{-1}\eta\right\}\right\},
\end{eqnarray}
with $\mathbf{A} = \text{diag}\left(\alpha_1,\alpha_2,\dots,\alpha_M\right)$. The Dirac delta function may be used as an approximation on the basis that this point-estimate is representative of the posterior in the sense that the integral calculation for the posterior using the point-estimate is roughly equal to the one obtained by sampling from the full posterior distribution \cite{tipping_sparse_2001}. This maximization is a type-\MakeUppercase{\romannumeral 2} maximum likelihood and can be calculated using a fast method \cite{tipping_fast_2003}. Next, TSBR integrates out $\alpha$ and $\sigma^2$ to get the posterior over the weight-vector:
\begin{eqnarray}\label{tsbr7}
  p\left(\mathbf{w}|\eta\right) &=& \iint p\left(\mathbf{w},\alpha,\sigma^2|\eta\right)d\alpha d\sigma^2   \nonumber \\
  &=& \iint p\left(\mathbf{w}|\eta,\alpha,\sigma^2\right)p\left(\alpha,\sigma^2|\eta\right)d\alpha d\sigma^2   \nonumber \\
  &\approx& \iint p\left(\mathbf{w}|\eta,\alpha,\sigma^2\right)\delta\left(\hat{\alpha}_\text{ML},\hat{\sigma}^2_\text{ML}\right)d\alpha d\sigma^2  \nonumber \\
  &=& p\left(\mathbf{w}|\eta,\hat{\alpha}_\text{ML},\hat{\sigma}^2_\text{ML}\right)    \nonumber \\
  &=& \frac{p\left(\eta|\mathbf{w},\hat{\sigma}^2_\text{ML}\right)p\left(\mathbf{w}|\hat{\alpha}_\text{ML}\right)}{p\left(\eta|\hat{\alpha}_\text{ML},\hat{\sigma}^2_\text{ML}\right)} \qquad\qquad\text{[Bayes' rule]}    \nonumber \\
  &=& \left(2\pi\right)^{-M/2}\left|\hat{\Sigma}\right|^{-1/2}\exp\left\{-\frac{1}{2}\left(\mathbf{w}-\hat{\mu}\right)^\text{T}\hat{\Sigma}^{-1}\left(\mathbf{w}-\hat{\mu}\right)\right\}  \nonumber \\
  &=& \mathcal{N}\left(\mathbf{w}|\hat{\mu},\hat{\Sigma}\right),
\end{eqnarray}
in which the posterior covariance and mean are:
\begin{eqnarray}
  \hat{\Sigma} &=& \left[\hat{\sigma}_\text{ML}^{-2}\Phi^\text{T}\Phi + \text{diag}\left(\hat{\alpha}_\text{ML}\right)\right]^{-1}  \label{tsbr8} \\
  \hat{\mu} &=& \hat{\sigma}_\text{ML}^{-2}\hat{\Sigma}\Phi^\text{T}\eta.    \label{tsbr9}
\end{eqnarray}
Therefore the posterior for each weight can be deduced from (\ref{tsbr7}):
\begin{equation}\label{tsbr10}
  p(w_j|\eta) = \mathcal{N}\left(w_j|\hat{\mu}_j,\hat{\Sigma}_{jj}\right),
\end{equation}
with mean $\hat{\mu}_j$ and standard deviation $\hat{\Sigma}_{jj}^{1/2}$. Thus the mean posterior prediction of the weight-vector $\mathbf{w}$ and other quantities that we want to obtain are determined by the values of $\eta$ and $\Phi$. This is the beauty of the Bayesian approach: there is no need to determine a regularization parameter via expensive cross-validation, and moreover likelihood values and confidence intervals for the solution can be easily calculated \cite{schmolck_smooth_2007}. Another merit of the Bayesian approach is that it can work with limited data since it incorporates prior information into the problem to supplement limited data.

\subsection{Implementation}
In practice we can show that the optimal values of many hyper-parameters $\alpha_j$ in (\ref{tsbr6}) are infinite \cite{tipping_sparse_2001} with further analysis in \cite{faul_analysis_2002, palmer_perspectives_2004}, and thus from (\ref{tsbr8})-(\ref{tsbr10}) the posterior distributions of many components of the weight-vector are sharply peaked at zero. From another perspective, the Bayesian framework is related to a series of re-weighted $L_1$ problems \cite{wipf_new_2008}. This leads to the sparsity of the resulting weight-vector. To further encourage the accuracy and robustness, a threshold $\delta \ge 0$ is placed on the model to filter out possible disturbance present in the weight-vector. Then the weight-vector is reestimated using the remaining terms, iteratively until convergence. The entire procedure is summarized in Algorithm \ref{algorithm_tsbr}. A discussion on how to choose the threshold and its impact on the solution is detailed in \cite{zhang_robust_2018}. As the threshold is a parameter representing the model complexity, we may use machine learning algorithms such as cross-validation to determine it. Note that in Algorithm \ref{algorithm_tsbr}, $\hat{\mu}$ is more and more sparse after each loop by design. Thus the convergence of the algorithm is guaranteed given the convergence of the calculation of maximum likelihood in (\ref{tsbr6}). TSBR follows the implementation in \cite{tipping_fast_2003} for the calculation of (\ref{tsbr6}), with the global convergence analysis in \cite{wipf_new_2008}.

Next, TSBR defines a model-selection criterion that quantifies the quality of the posterior estimated model as follows:
\begin{equation}\label{tsbr11}
  \text{Model-selection criterion} = \sum_{\substack{j=1 \\ \hat{\mu}_j \neq 0}}^{M}\frac{\hat{\Sigma}_{jj}}{\hat{\mu}_j^2},
\end{equation}
where each estimated variance $\hat{\Sigma}_{jj}$ is divided by the square of the corresponding estimated mean $\hat{\mu}_j^2$ to normalize the variance on each weight. This definition adds up all the normalized variances of each weight present in the result, and penalizes the unsureness of the estimations. In other words, a smaller model-selection criterion means smaller normalized variances and higher posterior confidence, and implies higher model quality. If given a set of candidate models for the data, the preferred model would be the one with the minimum model-selection criterion.

\begin{algorithm}[t]
\SetAlgoLined
\caption{Threshold sparse Bayesian regression: $\eta = \Phi \mathbf{w} + \epsilon$}
\label{algorithm_tsbr}
\KwIn{$\eta$, $\Phi$, threshold}
\KwOut{$\hat{\mu}$, $\hat{\Sigma}$}
  Calculate the posterior distribution $p\left(\mathbf{w}|\eta\right)$ in $\eta = \Phi \mathbf{w}$, and let the mean be $\hat{\mu}$\;
  For components of $\hat{\mu}$ with absolute value less than the threshold, set them as 0\;
  \While{$\hat{\mu}\neq \mathbf{0}$}{
    Delete the columns of $\Phi$ whose corresponding weight is 0, and let the result be $\Phi'$\;
    Calculate the posterior distribution $p\left(\mathbf{w}'|\eta\right)$ in $\eta = \Phi' \mathbf{w}'$, and let the mean be $\hat{\mu}'$\;
    Update the corresponding components of $\hat{\mu}$ using $\hat{\mu}'$\;
    For components of $\hat{\mu}$ with absolute value less than the threshold, set them as 0\;
    \If{$\hat{\mu}$ is the same as the one on the last loop}{
      break\;
    }
  }
  Set the submatrix of $\hat{\Sigma}$ corresponding to non-zero components of $\hat{\mu}$ as the last estimated posterior variance in the preceding procedure, and set the other elements of $\hat{\Sigma}$ as 0.
\end{algorithm}

\section{Subsampling-based threshold sparse Bayesian regression}
\label{sub}
This section introduces the novel subsampling-based algorithm based on the threshold sparse Bayesian regression algorithm reviewed in the last section.

\subsection{Motivation}
In the regression problem (\ref{intro3}), when we have more data than the number of unknown weights: $N>M$, we may use a subset of the data $\left\{t_i,x_i,x'_i\right\}_{i=1}^N$ to estimate the weights. We do this on the basis that the data sets collected from real-world cases may contain outliers or a percentage of data points of high noise. Classical methods for parameter estimation, such as least squares, fit a model to all of the presented data. These methods have no internal mechanism for detecting or discarding outliers. They are averaging methods based on the assumption (the smoothing assumption) that there will always be enough good values to smooth out any gross noise \cite{fischler_random_1981}. In practice the data may contain more noise than what the good values can ever compensate, breaking the smoothing assumption. To deal with this situation, an effective way is to discard the ``bad'' data points (outliers or those of high noise), and use the rest ``good'' data points to run estimations. Based on this idea, we propose an algorithm called subsampling-based threshold sparse Bayesian regression (SubTSBR). Our algorithm filters out bad data points by the means of random subsampling.

Other robust regression methods designed to overcome the limitations of traditional methods include iteratively reweighted least squares \cite{holland_robust_1977, chartrand_iteratively_2008}, random sample consensus \cite{fischler_random_1981}, least absolute deviations \cite{karst_linear_1958}, Theil-Sen estimator \cite{theil_rank-invariant_1992, sen_estimates_1968}, repeated median regression \cite{siegel_robust_1982, rousseeuw_least_1984}, and simultaneous variable selection and outlier identification \cite{hoeting_method_1996}. Also, the methods in \cite{farrell_protection_1994, box_bayesian_1968, west_outlier_1984} can handle outliers in the Bayesian framework through the design of the prior distribution. Recently, a method called entropic regression \cite{almomani_how_2020} is developed to deal with noise and outliers for the discovery of dynamics. There might be alternatives to our algorithm or ways to combine our algorithm with these methods.

\subsection{Implementation}
SubTSBR approaches the problem by selecting data points randomly to estimate the weights and using the model-selection criterion to evaluate the estimations. When an estimation is ``good'' (small model-selection criterion), our algorithm identifies the corresponding selected data points as good data points and the estimation as the final result. To be specific, our algorithm is given a user-preset subsampling size $S$ $(<N)$ and the number of subsamples $L$ $(\ge 1)$ at the very beginning. For each subsample, a subset of the data consisting of $S$ data points is randomly selected: $\left\{t_{k_i},x_{k_i},x'_{k_i}\right\}_{i=1}^S$ $\subset\left\{t_i,x_i,x'_i\right\}_{i=1}^N$ and used to estimate the weights $w_1$, $w_2$, \dots, $w_M$ in the following regression problem:
\begin{equation}\label{sub1}
    \left[
  \begin{array}{c}
    x_{k_1}' \\
    x_{k_2}' \\
    \vdots \\
    x_{k_S}'
  \end{array}
\right] = \left[
  \begin{array}{cccc}
    f_1(t_{k_1},x_{k_1}) & f_2(t_{k_1},x_{k_1}) & \cdots & f_M(t_{k_1},x_{k_1}) \\
    f_1(t_{k_2},x_{k_2}) & f_2(t_{k_2},x_{k_2}) & \cdots & f_M(t_{k_2},x_{k_2}) \\
    \vdots & \vdots & \ddots & \vdots \\
    f_1(t_{k_S},x_{k_S}) & f_2(t_{k_S},x_{k_S}) & \cdots & f_M(t_{k_S},x_{k_S}) \\
  \end{array}
\right]
\left[
  \begin{array}{c}
    w_1 \\
    w_2 \\
    \vdots \\
    w_M
  \end{array}
\right] + \epsilon,
\end{equation}
where $f_1(t,x),f_2(t,x),\dots,f_M(t,x)$ are the basis-functions and $\epsilon$ is the model error. This regression problem can be symbolized into the form as follows:
\begin{equation}\label{sub2}
  \eta = \Phi \mathbf{w} + \epsilon,
\end{equation}
which is (\ref{tsbr1}). By running TSBR (Algorithm \ref{algorithm_tsbr}), we obtain a differential equation:
\begin{equation}\label{sub3}
    \frac{dx}{dt} = \hat{\mu}_1 f_1(t,x) + \hat{\mu}_2 f_2(t,x) + \cdots + \hat{\mu}_M f_M(t,x),
\end{equation}
along with model-selection criterion calculated by (\ref{tsbr11}). After repeating this procedure $L$ times with different randomly selected data points, the differential equation with the smallest model-selection criterion among all the subsamples is chosen as the final result of the whole subsampling algorithm. Our algorithm has two user-preset parameters: the subsampling size and the number of subsamples. Their impact on the accuracy of the final result is discussed in Section \ref{example} through an example. Note that the above mechanism is described in the pattern (\ref{intro1}) for convenience and simplification. It also applies to higher-order differential equations and implicit differential equations, as long as the differential equations can be symbolized into the form (\ref{sub2}). The SubTSBR procedure is summarized in Algorithm \ref{algorithm_subtsbr}, where the for-loop can be coded parallelly.

\begin{algorithm}[t]
\SetAlgoLined
\caption{Subsampling-based threshold sparse Bayesian regression: $\eta = \Phi \mathbf{w} + \epsilon$}
\label{algorithm_subtsbr}
\KwIn{$\eta$, $\Phi$, threshold, subsampling size $S$, the number of subsamples $L$}
\KwOut{$\hat{\mu}$, $\hat{\Sigma}$}
  Let $I_{N\times N}$ be the $N\times N$ identity matrix, where $N$ is the number of rows in $\Phi$\;
  \For{$r = 1$ to $L$}{
    Let $P_r$ be an $S\times N$ submatrix of $I_{N\times N}$ with randomly chosen rows\;
    Use Algorithm \ref{algorithm_tsbr} to solve the problem $P_r\eta = P_r\Phi \mathbf{w} + \epsilon$, getting $\hat{\mu}_r, \hat{\Sigma}_r$\;
    Calculate $\text{[model-selection criterion]}_r$ using $\hat{\mu}_r, \hat{\Sigma}_r$ and (\ref{tsbr11})\;
  }
  Let $R=\arg\min_r \{\text{[model-selection criterion]}_r\}$\;
  Let $\hat{\mu}=\hat{\mu}_R$ and $\hat{\Sigma}=\hat{\Sigma}_R$.
\end{algorithm}

\subsection{Why it works}
The numerical results in this paper show that our subsampling algorithm can improve the overall accuracy in the discovery of differential equations, and the model-selection criterion (\ref{tsbr11}) is capable of evaluating the estimations. The given data $\left\{t_i,x_i,x'_i\right\}_{i=1}^N$ contain a part of data points of low noise and a part of data points of high noise. When a subset consisting of only data points of low noise is selected, our algorithm would estimate the weights well and indicate that this is the case by showing a small model-selection criterion (\ref{tsbr11}). As we do not know which data points are of low noise and which data points are of high noise before the model is discovered, we select a subset from the data randomly, repeating multiple times. When it happens that the selected data points are of low noise, we would have a good estimation of the weights and at the same time recognize this case.

The numerical results also show that when the subsampling size increases, the performance of our algorithm gets better and then worse. When the number of subsamples increases, the performance of our algorithm keeps getting better. In practice, as more subsamples mean more computational time, we can increase the number of subsamples adaptively and stop the algorithm when the smallest model-selection criterion among all the subsamples drops below a certain preset value or the smallest model-selection criterion stops decreasing.

\subsection{The number of subsamples needed to exclude outliers for a certain confidence level}
The outliers in the data can cause serious problems in the estimations and should be excluded. The subsampling procedure in Algorithm \ref{algorithm_subtsbr} is designed to be resistant to outliers. Here we calculate how many subsamples are needed to exclude the outliers from a data set for a certain confidence level.

Suppose we are given $N$ data points, a portion $p$ of which are outliers. Suppose the subsampling size is $S$. We try to determine the number of subsamples $L$ such that with confidence $q$, at least one of the $L$ randomly selected subsets of the data does not contain any outlier.

The number of outliers is $pN$ and the number of ``good'' data points is $(1-p)N$. For a random subset of size $S$ ($S \le (1-p)N$) not containing any outlier, the probability is
\begin{equation}
    \frac{{(1-p)N \choose S}}{{N \choose S}} = (1-p)\frac{(1-p)N-1}{N-1}\frac{(1-p)N-2}{N-2}\cdots \frac{(1-p)N-S+1}{N-S+1}.
\end{equation}
For a random subset of size $S$ containing at least one outlier, the probability is
\begin{equation}
    1 - \frac{{(1-p)N \choose S}}{{N \choose S}}.
\end{equation}
For $L$ random subsets of size $S$ each containing at least one outlier, the probability is
\begin{equation}
    \left[1 - \frac{{(1-p)N \choose S}}{{N \choose S}}\right]^L.
\end{equation}
For $L$ random subsets of size $S$ at least one subset not containing any outlier, the probability is
\begin{equation}
    1 - \left[1 - \frac{{(1-p)N \choose S}}{{N \choose S}}\right]^L.
\end{equation}
Set the probability equal to $q$:
\begin{equation}
    1 - \left[1 - \frac{{(1-p)N \choose S}}{{N \choose S}}\right]^L = q.
\end{equation}
Then we have
\begin{equation} \label{outlier1}
    L = \frac{\log (1-q)}{\log \left(1 - \frac{{(1-p)N \choose S}}{{N \choose S}}\right)}.
\end{equation}

The equation (\ref{outlier1}) gives the number of subsamples such that our algorithm selects at least one subsample without any outlier at the confidence level $q$. In practice, if we choose a larger number of subsamples, the confidence level goes up and we expect to get a better result, but the tradeoff is more computation time.

\section{Tackling high noise and outliers using subsampling-based threshold sparse Bayesian regression}
\label{example}
High noise and outliers in the data can hinder model-discovering algorithms from producing correct results. Through two examples, we demonstrate how to apply our algorithm to discover models. Also, we detail the mechanism of our algorithm and investigate the robustness against noise and outliers.

\subsection{Example: predator-prey model with noise}
The predator-prey model is a system of a pair of first-order nonlinear differential equations and is frequently used to describe the interaction between two species, one as a predator and the other as prey. The population change by time is as follows:
\begin{eqnarray}
\frac{dx}{dt} &=& \alpha x - \beta x y \label{ex1} \\
\frac{dy}{dt} &=& \delta x y - \gamma y \label{ex2},
\end{eqnarray}
where $x$ is the number of the prey, $y$ is the number of the predator, and $\alpha, \beta, \delta, \gamma$ are positive real parameters describing the interaction of the two species. In this example, we fix the parameters as follows:
\begin{eqnarray}
\frac{dx}{dt} &=& \frac{1}{2} x - \frac{3}{2} x y \label{ex3} \\
\frac{dy}{dt} &=& x y - \frac{1}{2} y. \label{ex4}
\end{eqnarray}
We assume that we do not know about the formulas for the system (\ref{ex3}) - (\ref{ex4}), neither the terms nor the parameters, and try to discover the model using noisy data.

\subsubsection{Data collection}
\begin{figure}[t]\centering
  (a)\subfloat{\includegraphics[width=7cm]{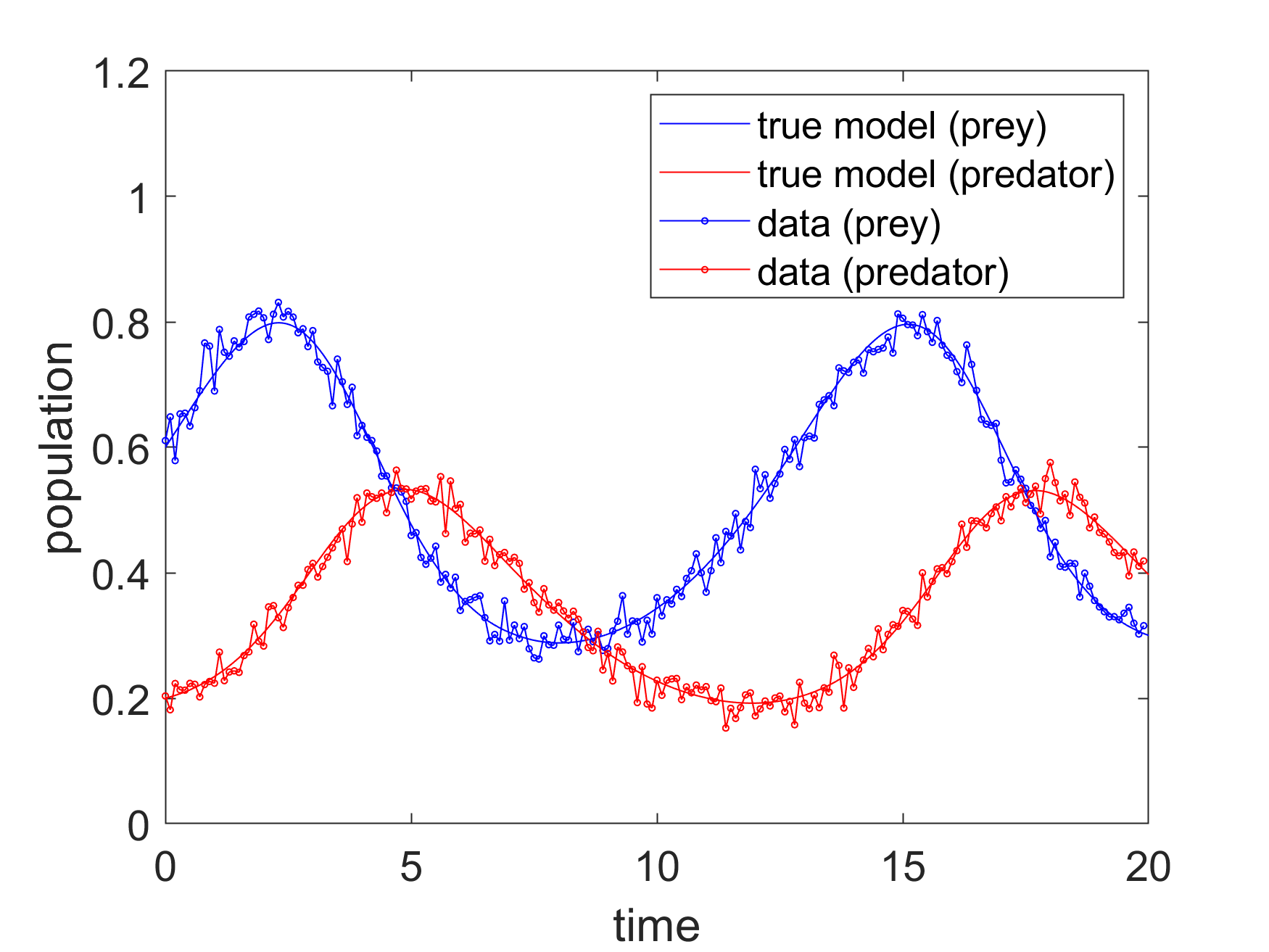}\label{ex5a}}
  (b)\subfloat{\includegraphics[width=7cm]{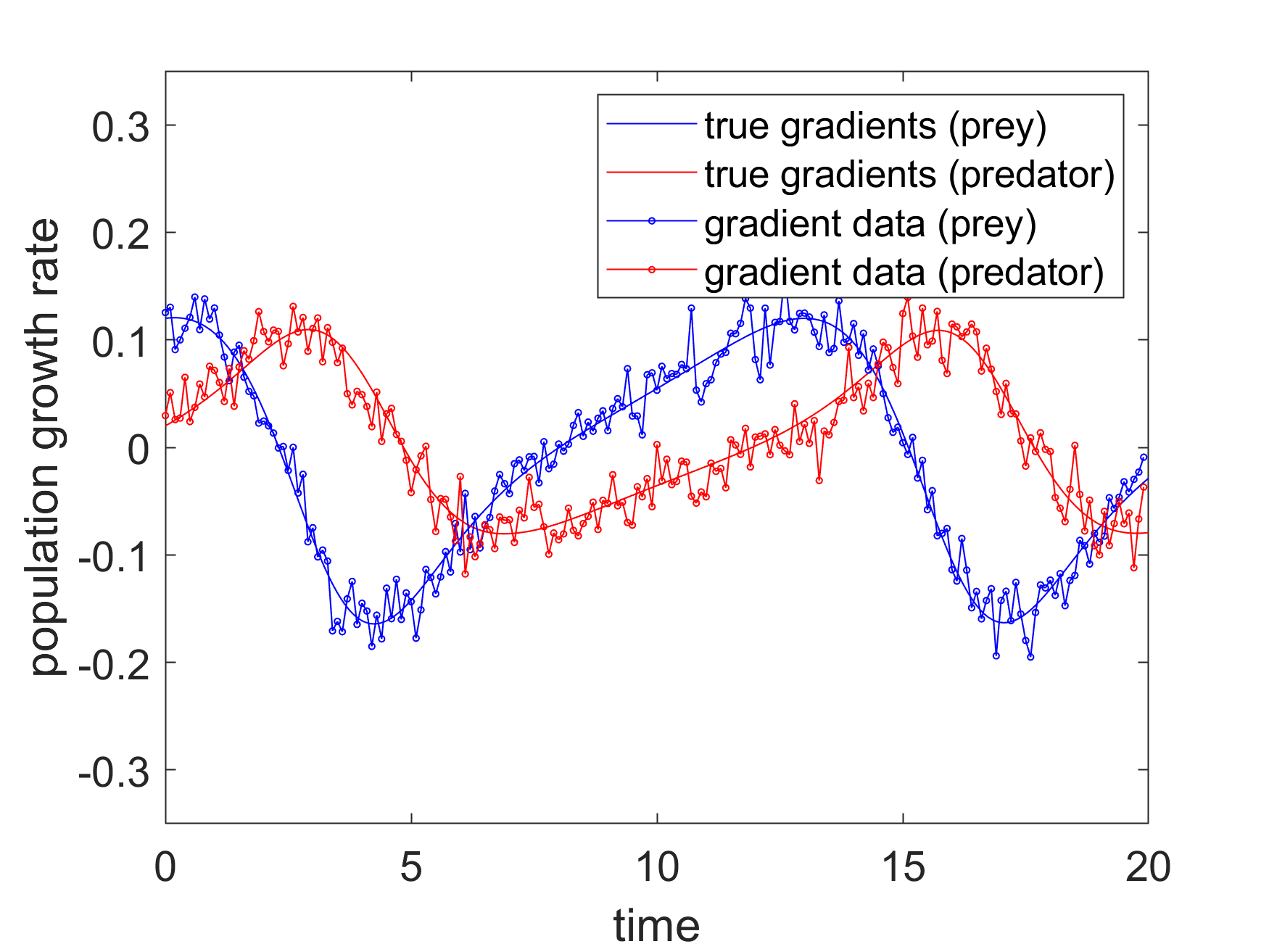}\label{ex5b}}
  \caption{[Predator-prey model] (a) The true predator-prey model and the noisy data. (b) The true gradients and the noisy gradient data.}
  \label{ex5}
\end{figure}

We first generate $200$ data points with gradients from the system (\ref{ex3}) - (\ref{ex4}), with the initial value $x_0=0.6$ and $y_0=0.2$, during time $t=0$ to $t=20$. Then independent and identically distributed white noise $\mathcal{N}(0,0.02^2)$ is added to all the data $x$, $y$ and all the gradients $dx/dt$, $dy/dt$. See Figure \ref{ex5a} for the true model and the noisy data. See Figure \ref{ex5b} for the true gradients and the noisy gradient data. If we define the signal-to-noise ratio as
\begin{equation}\label{ex5-}
  \text{SNR}_{\text{dB}} = 10 \log_{10}\left(\frac{P_{\text{signal}}}{P_{\text{noise}}}\right) = 10 \log_{10}\left(\frac{||\text{signal}||_2^2}{||\text{noise}||_2^2}\right),
\end{equation}
then the $\text{SNR}_{\text{dB}}$ for $x$, $y$, $dx/dt$, $dy/dt$ are $28.3$, $25.7$, $14.5$, $10.2$, respectively.

\subsubsection{Discover the model using different sparse algorithms}
Now we try to discover the predator-prey model (\ref{ex3}) - (\ref{ex4}) using threshold sparse Bayesian regression (TSBR, Algorithm \ref{algorithm_tsbr}) and subsampling-based threshold sparse Bayesian regression (SubTSBR, Algorithm \ref{algorithm_subtsbr}). The basis-functions are monomials generated by $\{1, x, y\}$ up to degree three ($10$ terms in total). Both TSBR and SubTSBR have the threshold set at $0.1$. In addition, SubTSBR has the subsampling size set at $60$ and the number of subsamples set at $30$. TSBR uses all $200$ data points at one time to discover the model while SubTSBR does subsampling.

\begin{table}[p]
\begin{tabular}{|c|r|r|r|r|r|}
  \hline
  Term    & \multicolumn{1}{c|}{True} & \multicolumn{1}{c|}{TSBR} & \multicolumn{1}{c|}{SubTSBR} \\ \hline
  $1$     &        &                  &                  \\ \hline
  $x$     & $0.5$  & $0.285 (0.010)$  & $0.499 (0.016)$  \\ \hline
  $y$     &        &                  &                  \\ \hline
  $x^2$   &        &                  &                  \\ \hline
  $xy$    & $-1.5$ &                  & $-1.491 (0.044)$ \\ \hline
  $y^2$   &        & $-0.413 (0.119)$ &                  \\ \hline
  $x^3$   &        &                  &                  \\ \hline
  $x^2 y$ &        &                  &                  \\ \hline
  $x y^2$ &        & $-2.336 (0.096)$ &                  \\ \hline
  $y^3$   &        & $1.020 (0.241)$  &                  \\ \hline
\end{tabular}
\caption{[Predator-prey model] The true model for $dx/dt$ (\ref{ex3}) and the models discovered by threshold sparse Bayesian regression (TSBR) and subsampling-based threshold sparse Bayesian regression (SubTSBR). Every column represents the weights for the terms in the model. Blank means the model does not have the specific term. The numbers inside the parentheses following the weights in TSBR and SubTSBR represent the standard deviation for the weight.} \label{ex8}
\end{table}

\begin{table}[p]
\begin{tabular}{|c|r|r|r|r|r|}
  \hline
  Term    & \multicolumn{1}{c|}{True} & \multicolumn{1}{c|}{TSBR} & \multicolumn{1}{c|}{SubTSBR} \\ \hline
  $1$     &        &                  &                  \\ \hline
  $x$     &        &                  &                  \\ \hline
  $y$     & $-0.5$ & $-0.285 (0.013)$ & $-0.484 (0.020)$ \\ \hline
  $x^2$   &        &                  &                  \\ \hline
  $xy$    & $1$    &                  & $0.955 (0.038)$  \\ \hline
  $y^2$   &        &                  &                  \\ \hline
  $x^3$   &        &                  &                  \\ \hline
  $x^2 y$ &        & $0.909 (0.022)$  &                  \\ \hline
  $x y^2$ &        &                  &                  \\ \hline
  $y^3$   &        & $0.150 (0.048)$  &                  \\ \hline
\end{tabular}
\caption{[Predator-prey model] The true model for $dy/dt$ (\ref{ex4}) and the discovered models. All other interpretations are the same as Table \ref{ex8}.} \label{ex9}
\end{table}

\begin{figure}[t]\centering
  (a)\subfloat{\includegraphics[width=7cm]{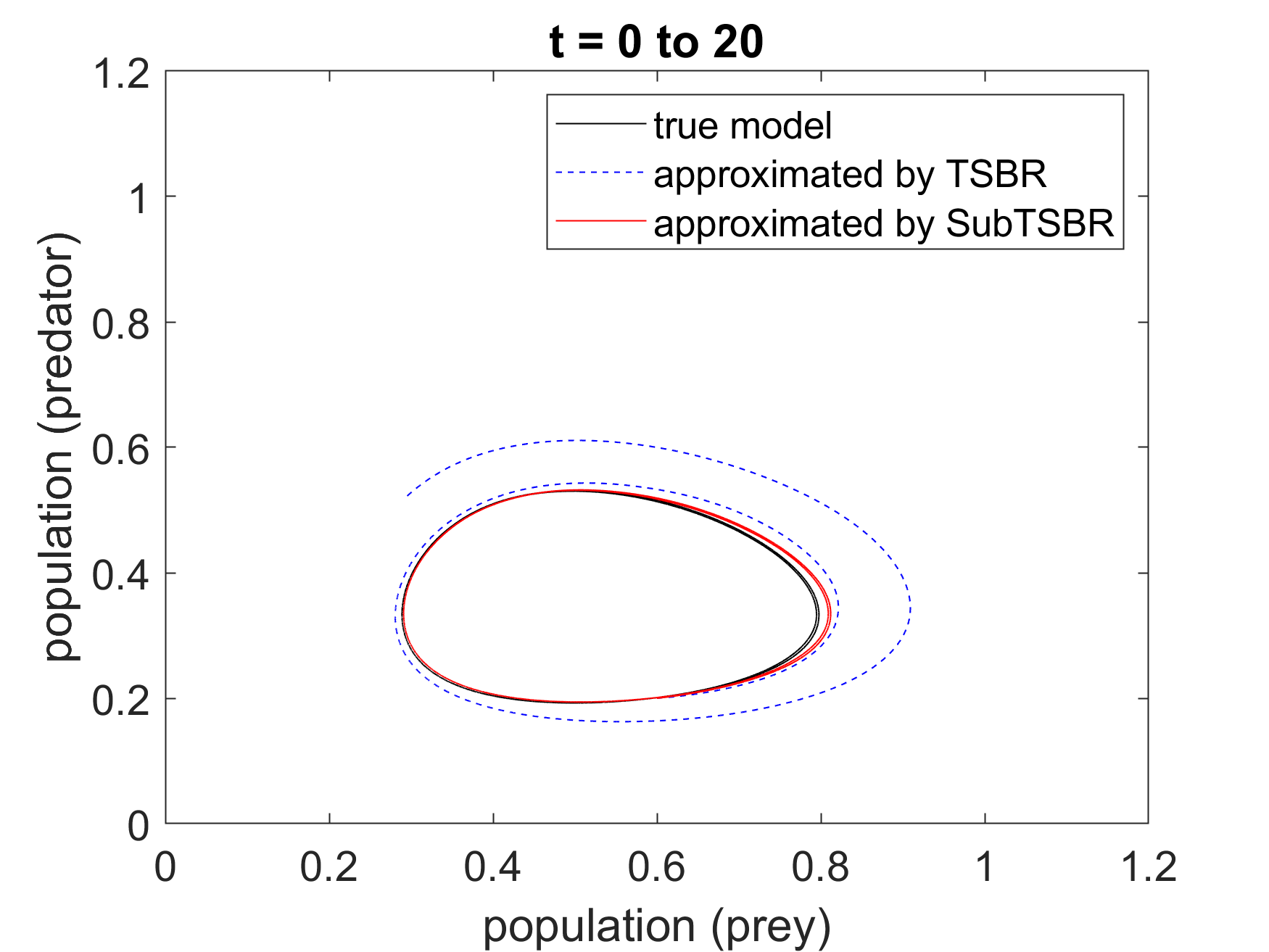}\label{ex12a}}
  (b)\subfloat{\includegraphics[width=7cm]{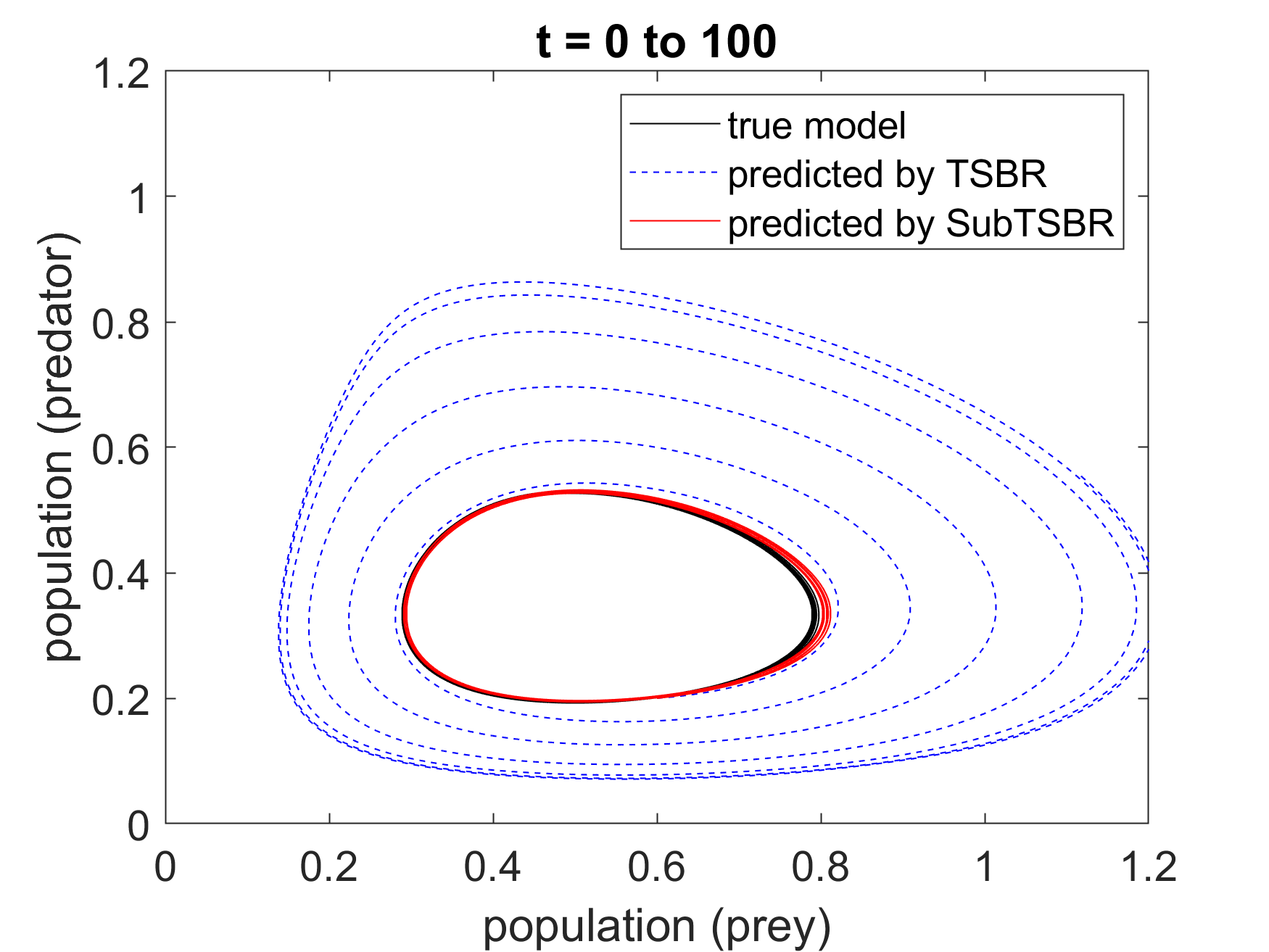}\label{ex12b}}
  \caption{[Predator-prey model] (a) Approximated dynamics by TSBR and SubTSBR from $t=0$ to $t=20$. (b) Predicted dynamics from $t=0$ to $t=100$.}
  \label{ex12}
\end{figure}

The numerical results are listed in Table \ref{ex8} and Table \ref{ex9}, from which we can see that SubTSBR approximates the true model better than TSBR. Although the data contain a considerable amount of noise, SubTSBR successfully finds the exact terms in the true model and accurately estimates the parameters. See Figure \ref{ex12} for the dynamics calculated by TSBR and SubTSBR. Note that since the data are collected from $t=0$ to $t=20$, Figure \ref{ex12a} shows the approximation and Figure \ref{ex12b} shows the prediction. SubTSBR demonstrates better performance than TSBR especially in prediction.

\subsubsection{Basis-selection success rate vs subsampling size and the number of subsamples}
\begin{figure}[t]\centering
  (a)\subfloat{\includegraphics[width=7cm]{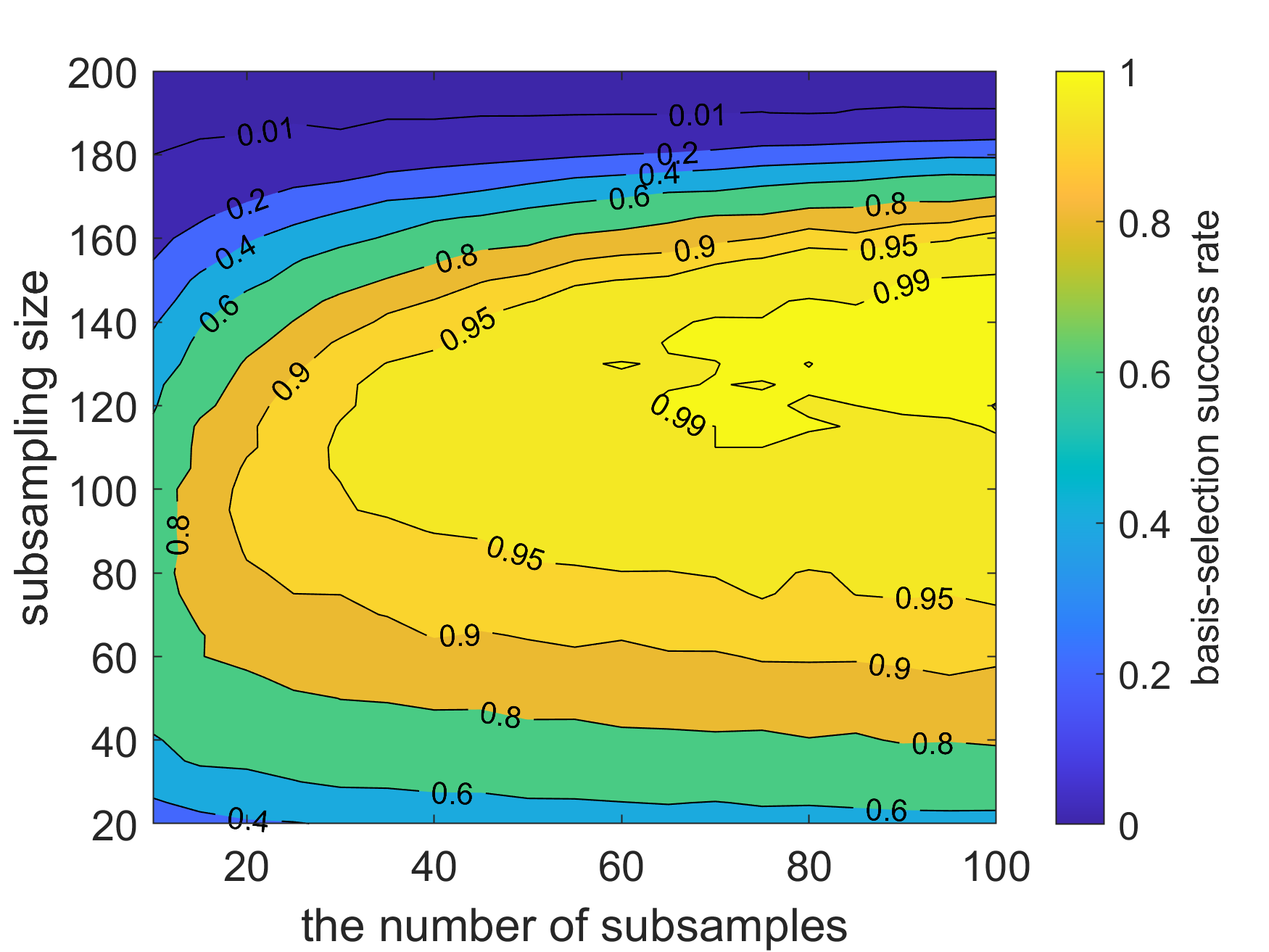}\label{ex13a}}
  (b)\subfloat{\includegraphics[width=7cm]{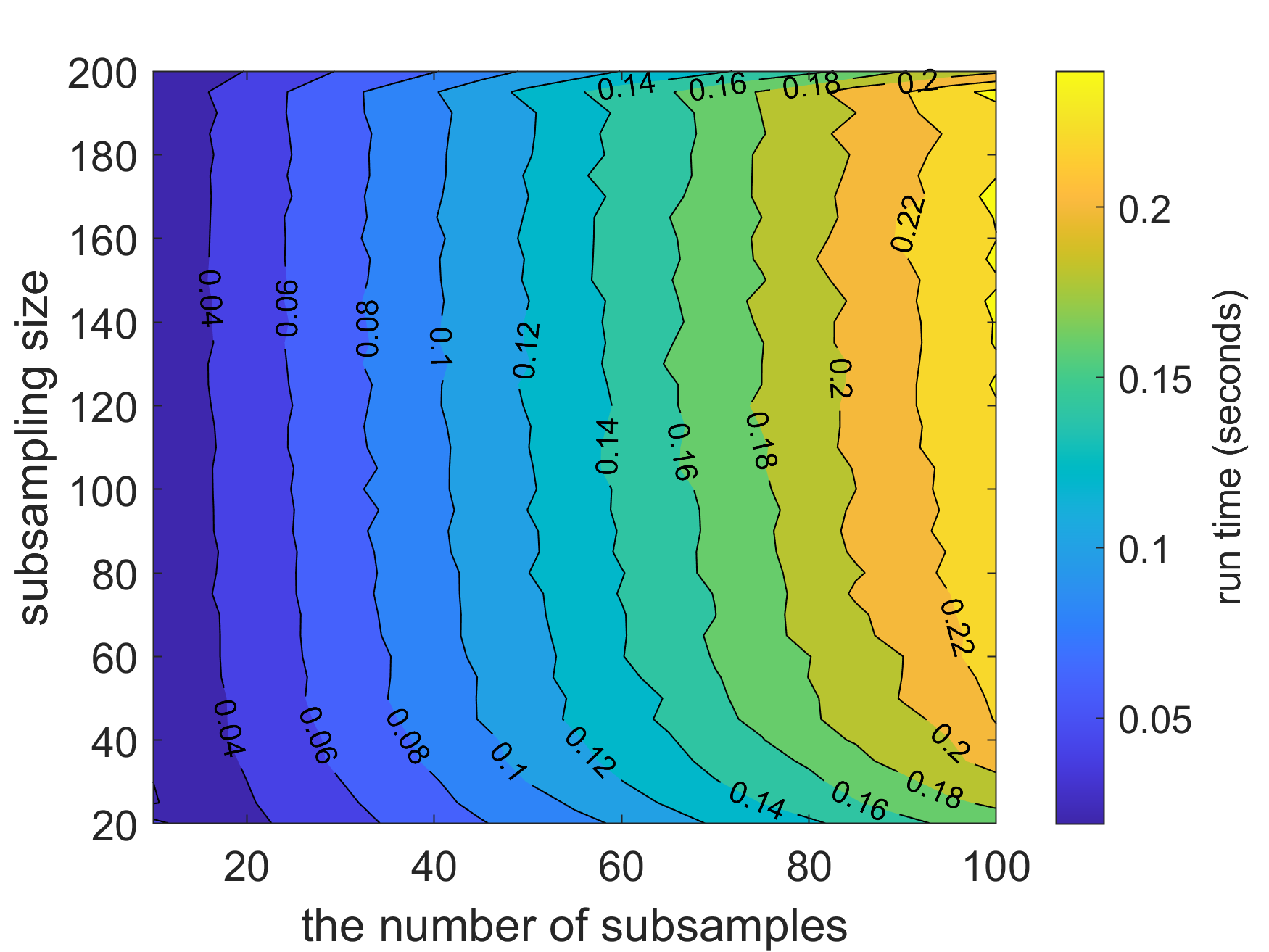}\label{ex13b}} \\
  (c)\subfloat{\includegraphics[width=7cm]{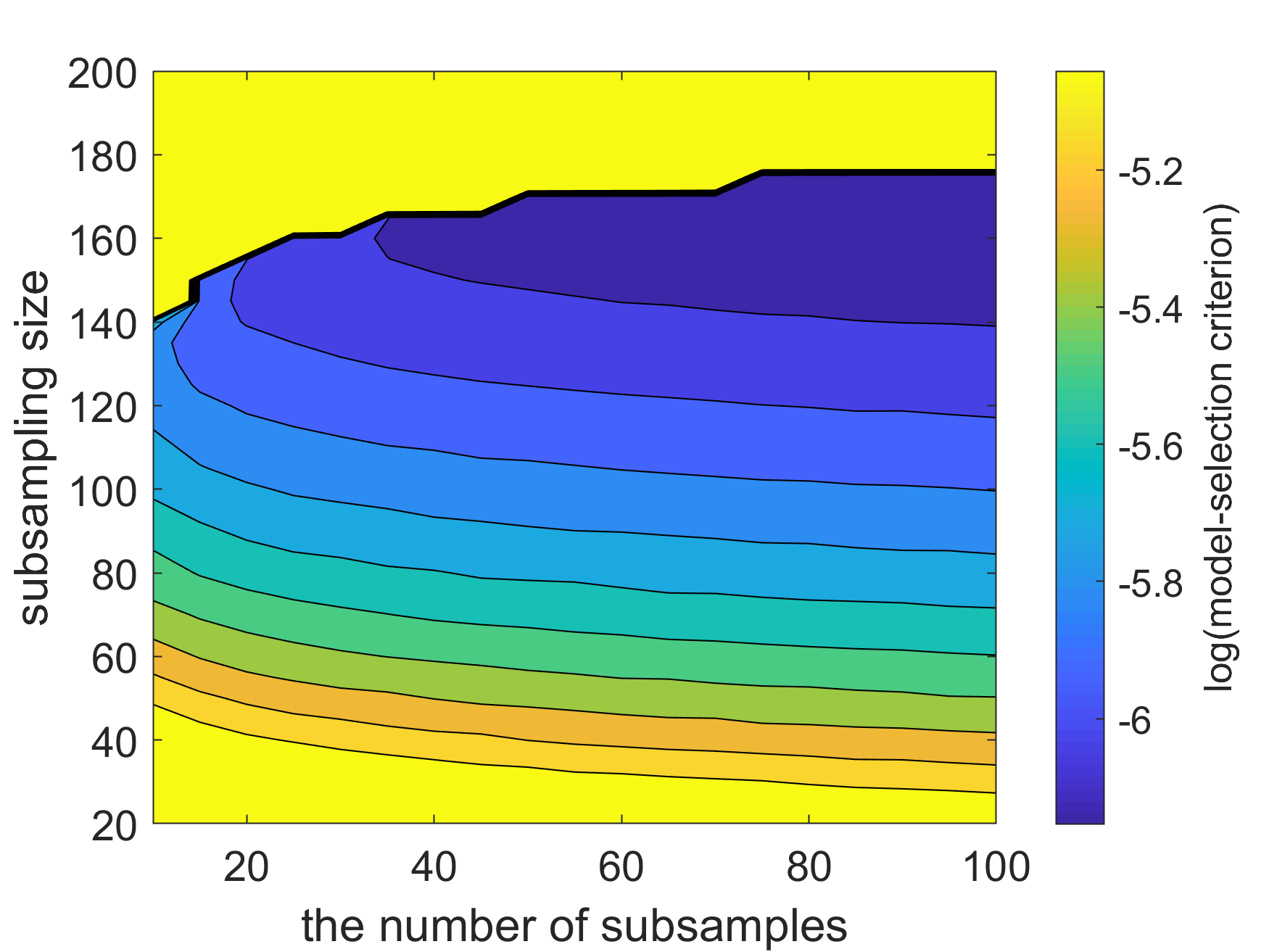}\label{ex13c}}
  (d)\subfloat{\includegraphics[width=7cm]{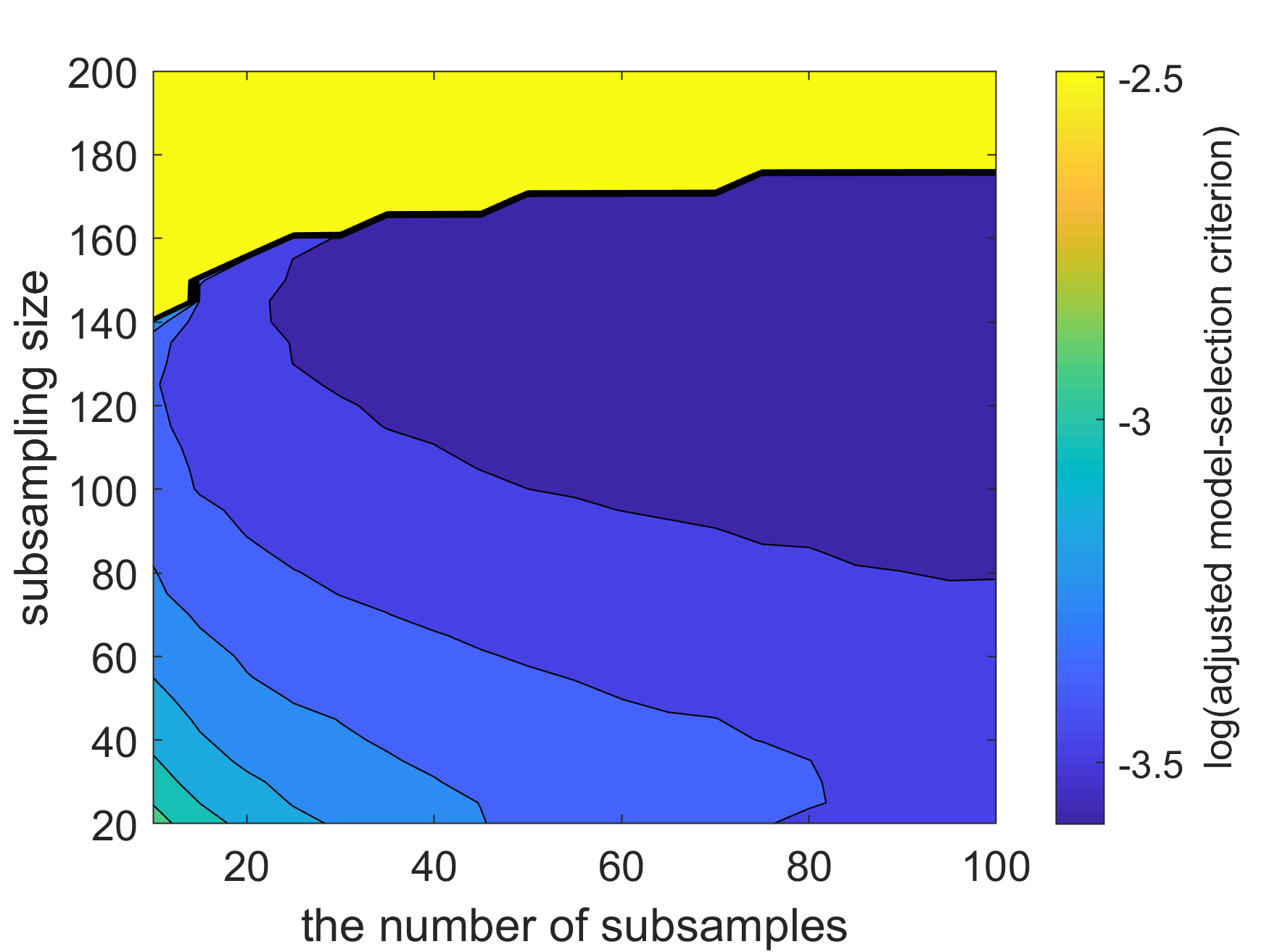}\label{ex13d}}
  \caption{[Predator-prey model] The total number of data points is $200$. (a) Basis-selection success rate vs subsampling size and the number of subsamples. (b) Computation time vs subsampling size and the number of subsamples. (c) Model-selection criterion vs subsampling size and the number of subsamples. (d) Adjusted model-selection criterion vs subsampling size and the number of subsamples.}
  \label{ex13}
\end{figure}

In the subsampling process of SubTSBR (Algorithm \ref{algorithm_subtsbr}), we have two parameters to set, one is the subsampling size and the other is the number of subsamples. Here, we investigate the impact on the basis-selection success rate by different subsampling sizes and different numbers of subsamples. In Figure \ref{ex13a}, for each subsampling size and each number of subsamples, SubTSBR is applied to the same data set collected above for $2000$ times. Then the percentage of successful identification of the exact terms in the system (\ref{ex3}) - (\ref{ex4}) is calculated and plotted.

For each fixed number of subsamples, basis-selection success rate goes up and then down when the subsampling size increases. When the subsampling size equals $200$, all the data points are used and SubTSBR is equivalent to TSBR (Algorithm \ref{algorithm_tsbr}). In this case the true terms cannot be identified. In addition, for each chosen number of subsamples there is an optimal subsampling size, and the optimal subsampling size increases as the number of subsamples increases.

For each fixed subsampling size, basis-selection success rate keeps going up when the number of subsamples increases. In addition, the contour lines are denser at larger subsampling size, which indicates that the basis-selection success rate benefits more from the increase of the number of subsamples when the subsampling size is larger.

This is because our data set is polluted by Gaussian noise and naturally contains some data points of high noise and some of low noise. As the subsampling size gets bigger, it is less likely for each of the random subsets of data to exclude the data points of high noise. When more subsamples are used, the likelihood for one of the subsamples to exclude the data points of high noise increases. As long as one of the subsamples excludes the data points of high noise, this subsample may successfully select the true basis functions and have the smallest model-selection criterion. When it happens, the final result would come from this subsample and SubTSBR selects the true basis functions successfully. On the other hand, when more data points are used in a subsample, the noise inside the subsample gets smoothed out easier in the regression (\ref{sub2}). If all the included data points in the subsample are of low noise, then the result from a larger subsampling size is more likely to be a better one. In conclusion, there is tradeoff for larger subsampling size---it is more difficult to include only data points of low noise while it is easier to smooth out the noise inside the subsample.

\subsubsection{Computation time vs subsampling size and the number of subsamples}
In Figure \ref{ex13b}, for each subsampling size and each number of subsamples, SubTSBR is applied to the same data set collected above to discover the system (\ref{ex3}) - (\ref{ex4}) for $2000$ times. Then the average computation time for each run is plotted. We can see that the computation time is in proportion to the number of subsamples.

\subsubsection{Adjusted model-selection criterion and auto-fitting of subsampling size}
In real-world applications, the situation is usually more complicated and the problem of setting the best subsampling size is subtle. Since the true equations are unknown, the basis-selection success rate cannot be calculated. Therefore, drawing a contour map like Figure \ref{ex13a} to find the best subsampling size is not available. Here we define the adjusted model-selection criterion as an indicator for the quality of the approximated model and fit the subsampling size automatically.

The model-selection criterion defined in (\ref{tsbr11}) depends on the number of data points used and the quality of the approximated model. The $\hat{\Sigma}_{jj}$ in (\ref{tsbr11}) is calculated by (\ref{tsbr8}). When the number of data points increases, the number of rows in the matrix $\Phi$ increases, causing a decrease of $\hat{\Sigma}_{jj}$. As a result, when the subsampling size is close to optimal, the model-selection criterion is dominated by and negatively correlated with the subsampling size. If we want to compare the quality of the results among different subsampling sizes, we need to adjust the model-selection criterion such that it is not directly related to the subsampling size and it depends solely on the quality of the model. Here we give the following empirical formula:
\begin{equation} \label{ex14}
    \text{Adjusted model-selection criterion} = \text{[model-selection criterion]} \times \text{[subsampling size]}^{0.5}.
\end{equation}

Now we use the example in this section to validate the formula (\ref{ex14}). In Figure \ref{ex13c} or Figure \ref{ex13d}, for each subsampling size and each number of subsamples, we run SubTSBR for $2000$ times and discover $2000$ models with their model-selection criterion or adjusted model-selection criterion. Then the logarithm of the median of the $2000$ model-selection criterion or adjusted model-selection criterion is plotted. Comparing Figure \ref{ex13c} with Figure \ref{ex13d}, we see that the adjusted model-selection criterion prefers models with smaller subsampling size. By choosing a model with the smallest adjusted model-selection criterion, the algorithm tends to find a model with the subsampling size in the region that produces the highest basis-selection success rate in Figure \ref{ex13a}. In this way, the subsampling size is fitted automatically by the algorithm. We do not have to set the subsampling size at the beginning but we may try different subsampling sizes to discover the model, and the best result can be selected from all the results.

\subsubsection{A new data set with lower noise}
\begin{figure}[p]\centering
  (a)\subfloat{\includegraphics[width=7cm]{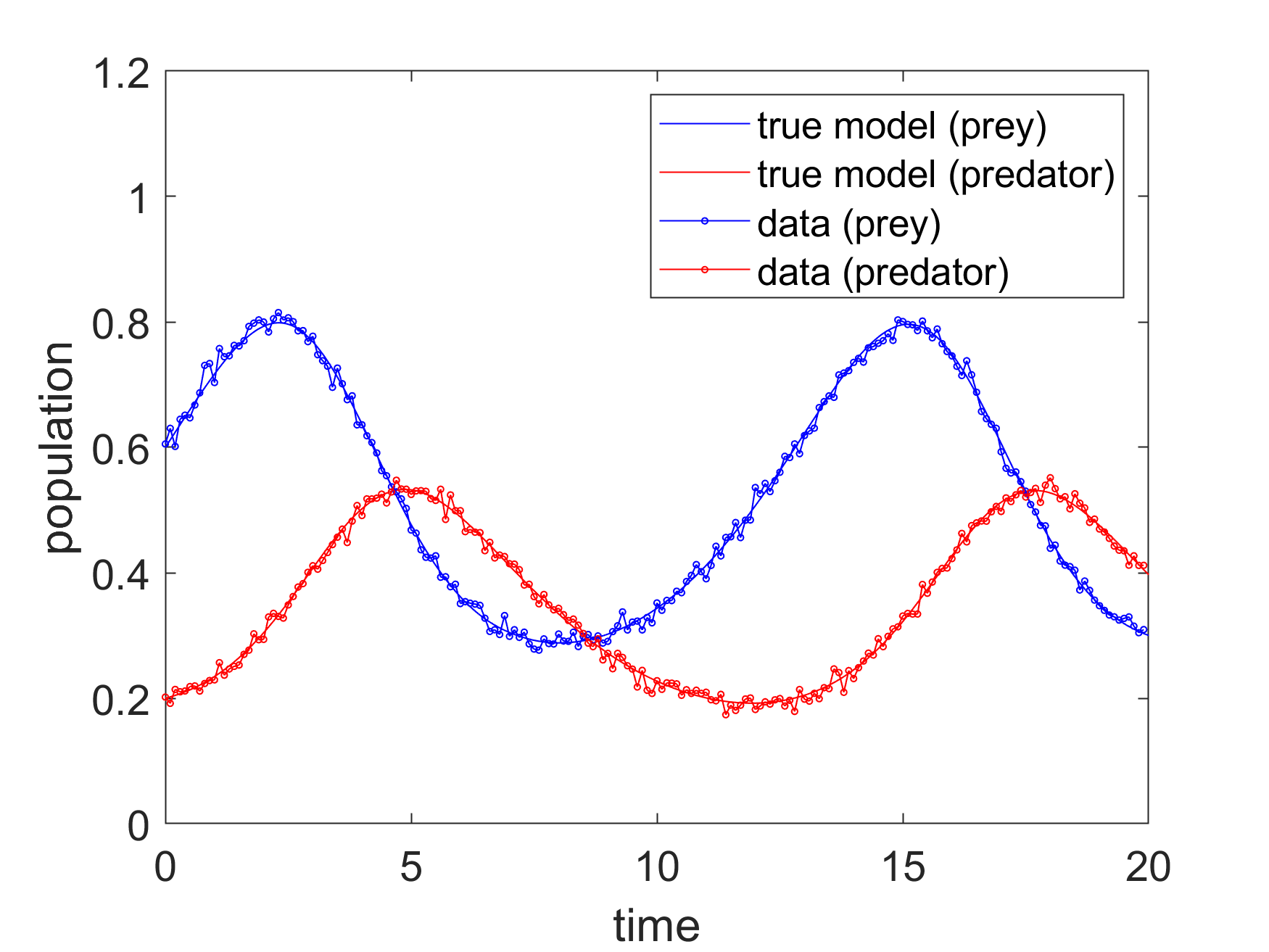}\label{ex17a}}
  (b)\subfloat{\includegraphics[width=7cm]{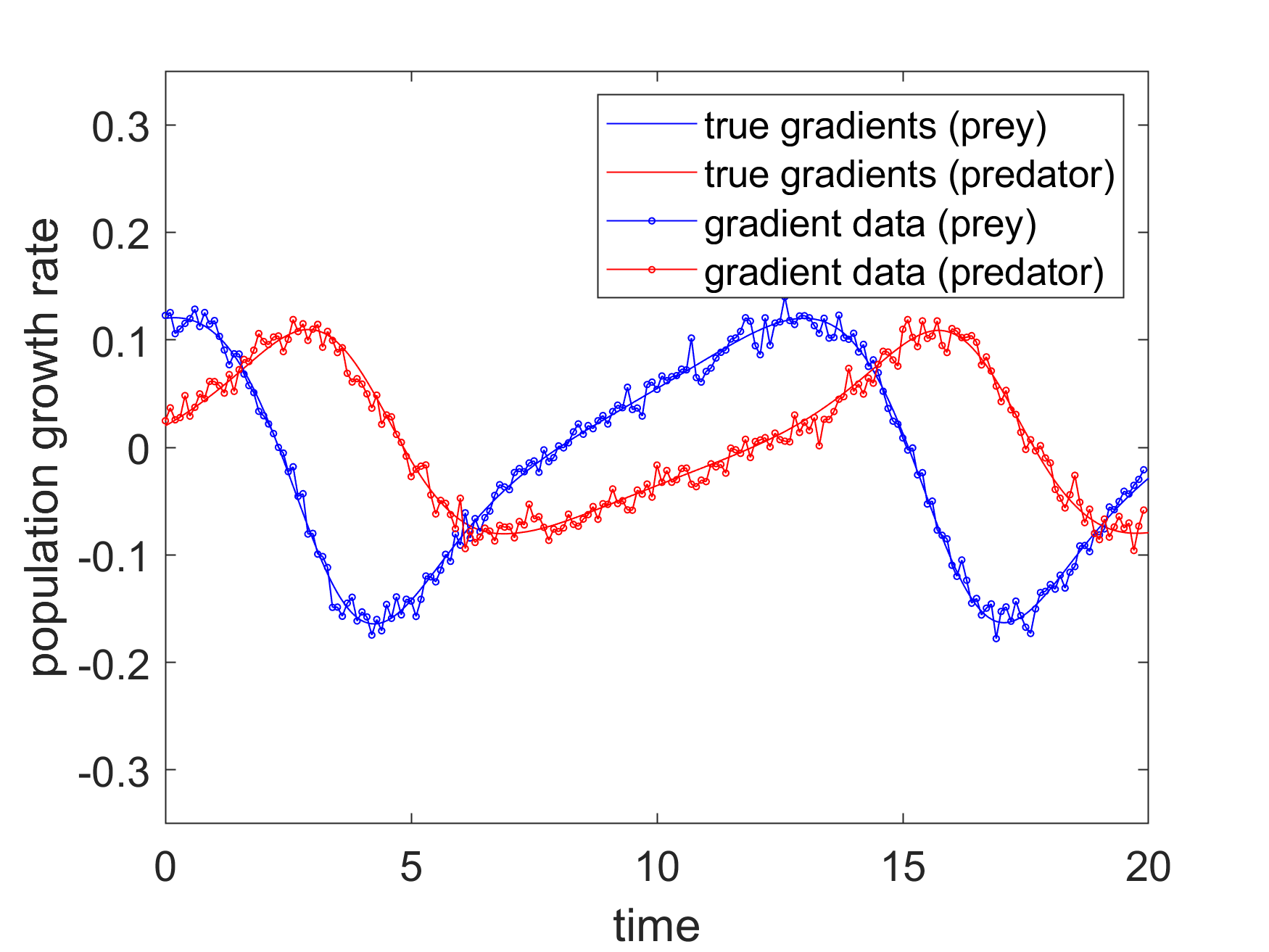}\label{ex17b}} \\
  (c)\subfloat{\includegraphics[width=7cm]{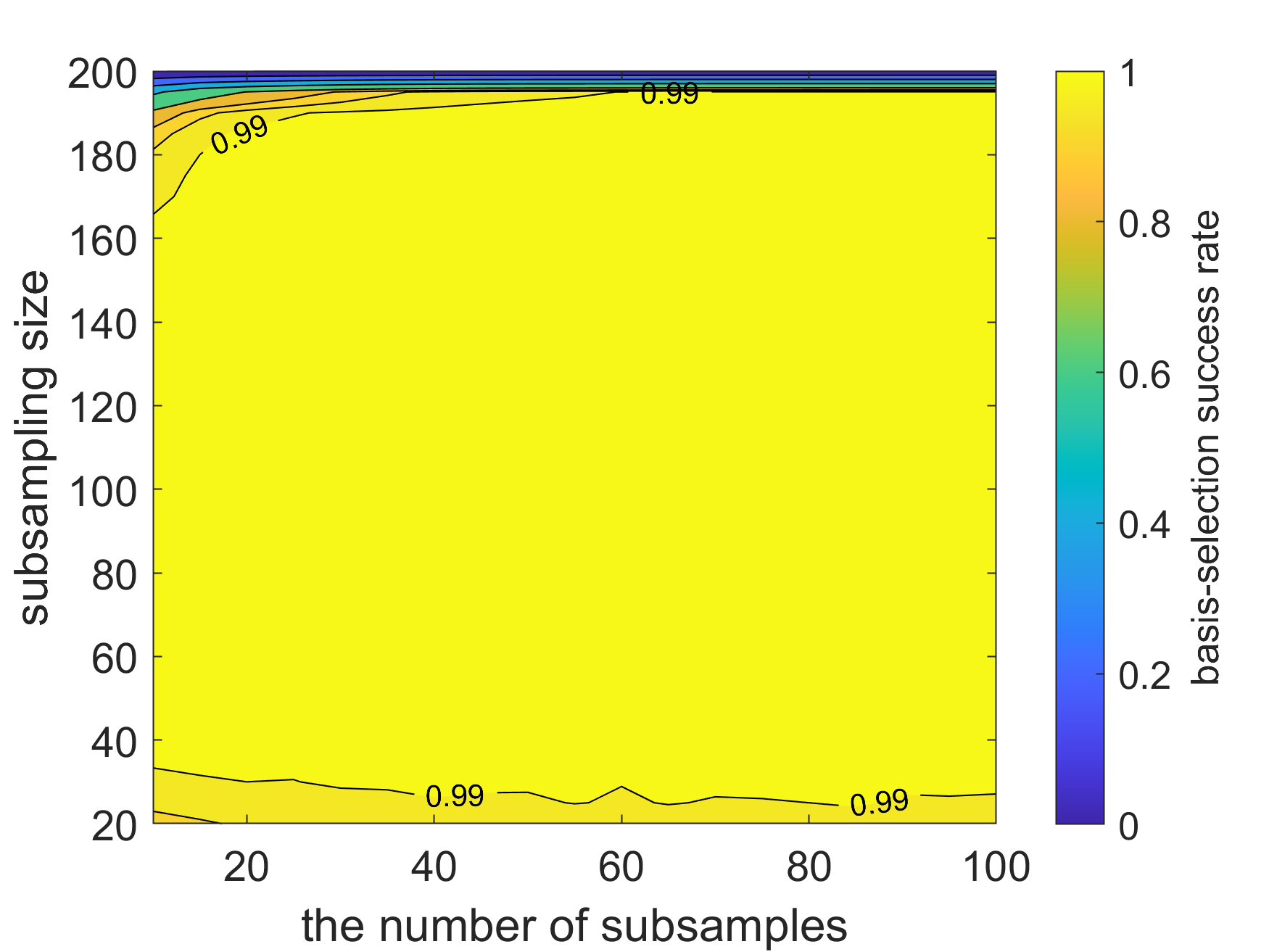}\label{ex17c}}
  (d)\subfloat{\includegraphics[width=7cm]{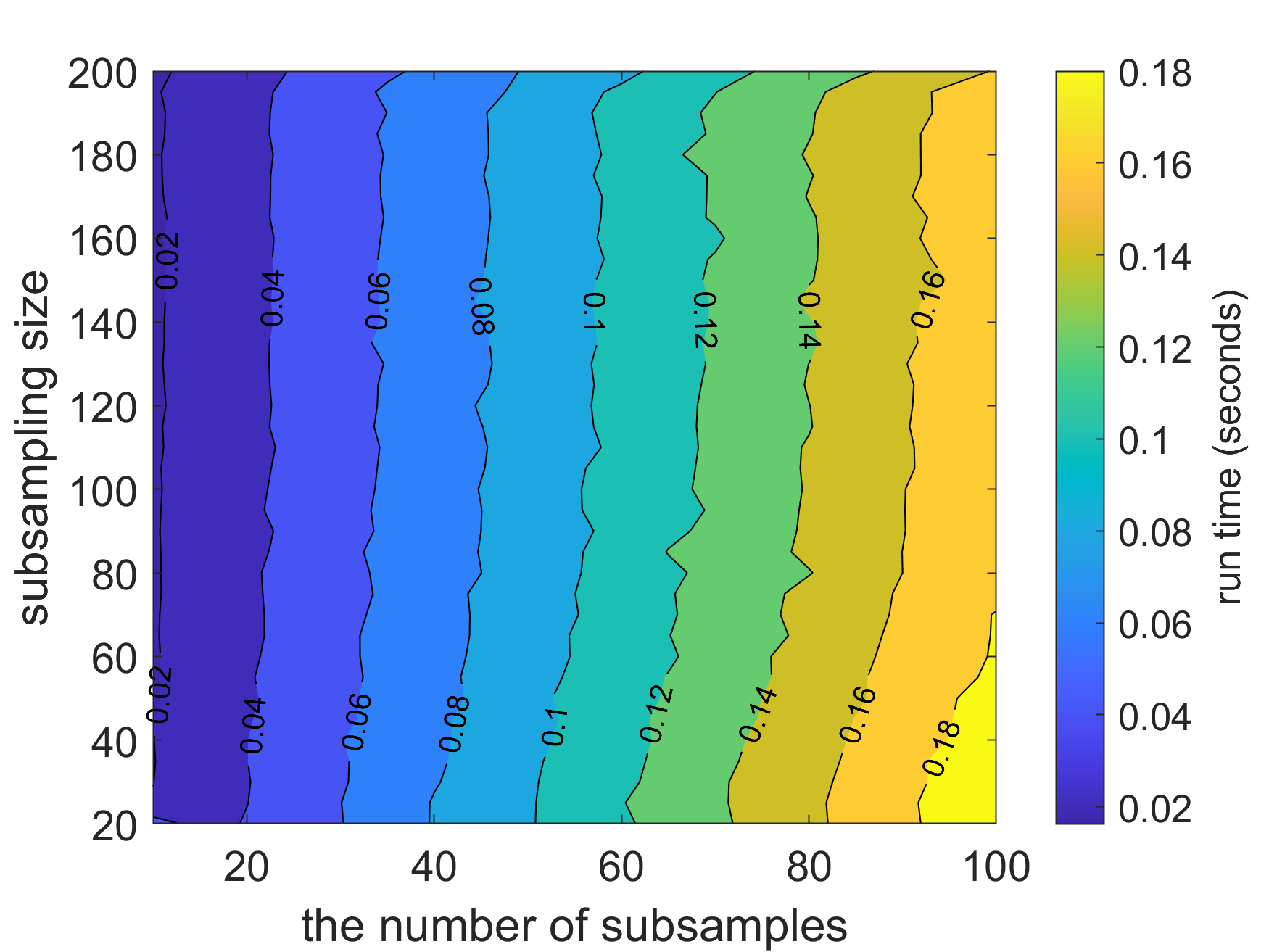}\label{ex17d}} \\
  (e)\subfloat{\includegraphics[width=7cm]{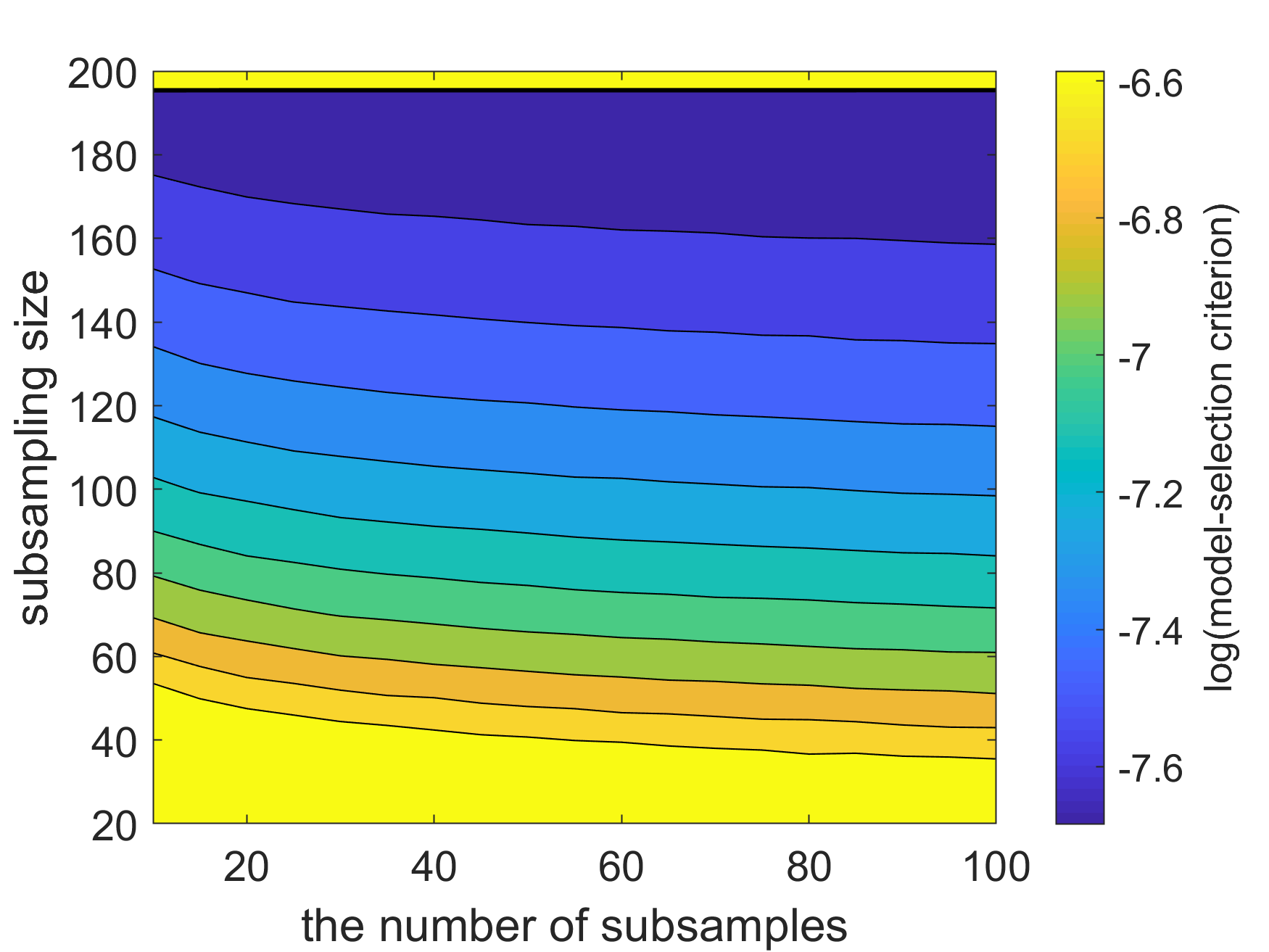}\label{ex17e}}
  (f)\subfloat{\includegraphics[width=7cm]{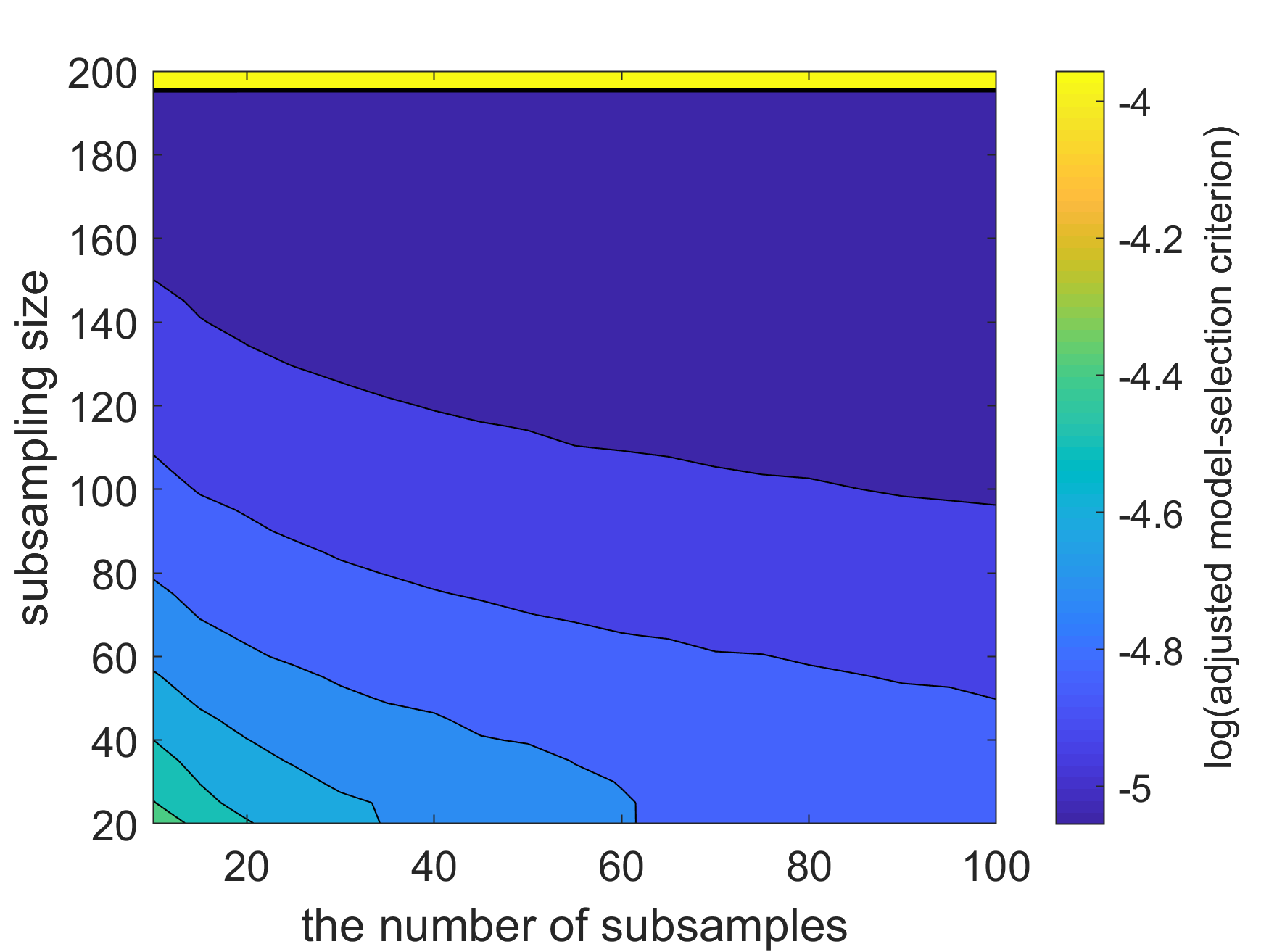}\label{ex17f}}
  \caption{[Predator-prey model] (a) The noisy data with lower noise than Figure \ref{ex5a}. (b) The noisy gradient data with lower noise than Figure \ref{ex5b}. (c) The basis-selection success rate with lower noise than Figure \ref{ex13a}. (d) The computation time with lower noise than Figure \ref{ex13b}. (e) The model-selection criterion with lower noise than Figure \ref{ex13c}. (f) The adjusted model-selection criterion with lower noise than Figure \ref{ex13d}.}
  \label{ex17}
\end{figure}

We use a new data set with white noise $\mathcal{N}(0,0.01^2)$ to discover the predator-prey model (\ref{ex3}) - (\ref{ex4}). The noise here is lower than the noise in the previous data set ($\mathcal{N}(0,0.02^2)$). All other settings remain the same. Corresponding to Figure \ref{ex5}, the noisy data and the noisy gradient data are presented in Figure \ref{ex17a} and Figure \ref{ex17b}. By the definition (\ref{ex5-}), the $\text{SNR}_{\text{dB}}$ for $x$, $y$, $dx/dt$, $dy/dt$ are $34.3$, $31.7$, $20.5$, $16.2$, respectively. Corresponding to Figure \ref{ex13}, the basis-selection success rate is presented in Figure \ref{ex17c}, the computation time is presented in Figure \ref{ex17d}, the model-selection criterion is presented in Figure \ref{ex17e}, and the adjusted model-selection criterion is presented in Figure \ref{ex17f}. With lower noise, the chance of successfully picking out the true terms from the basis-functions is higher.

\subsubsection{Error within a subsample vs model-selection criterion}
\begin{figure}[t]\centering
  (a)\subfloat{\includegraphics[width=7cm]{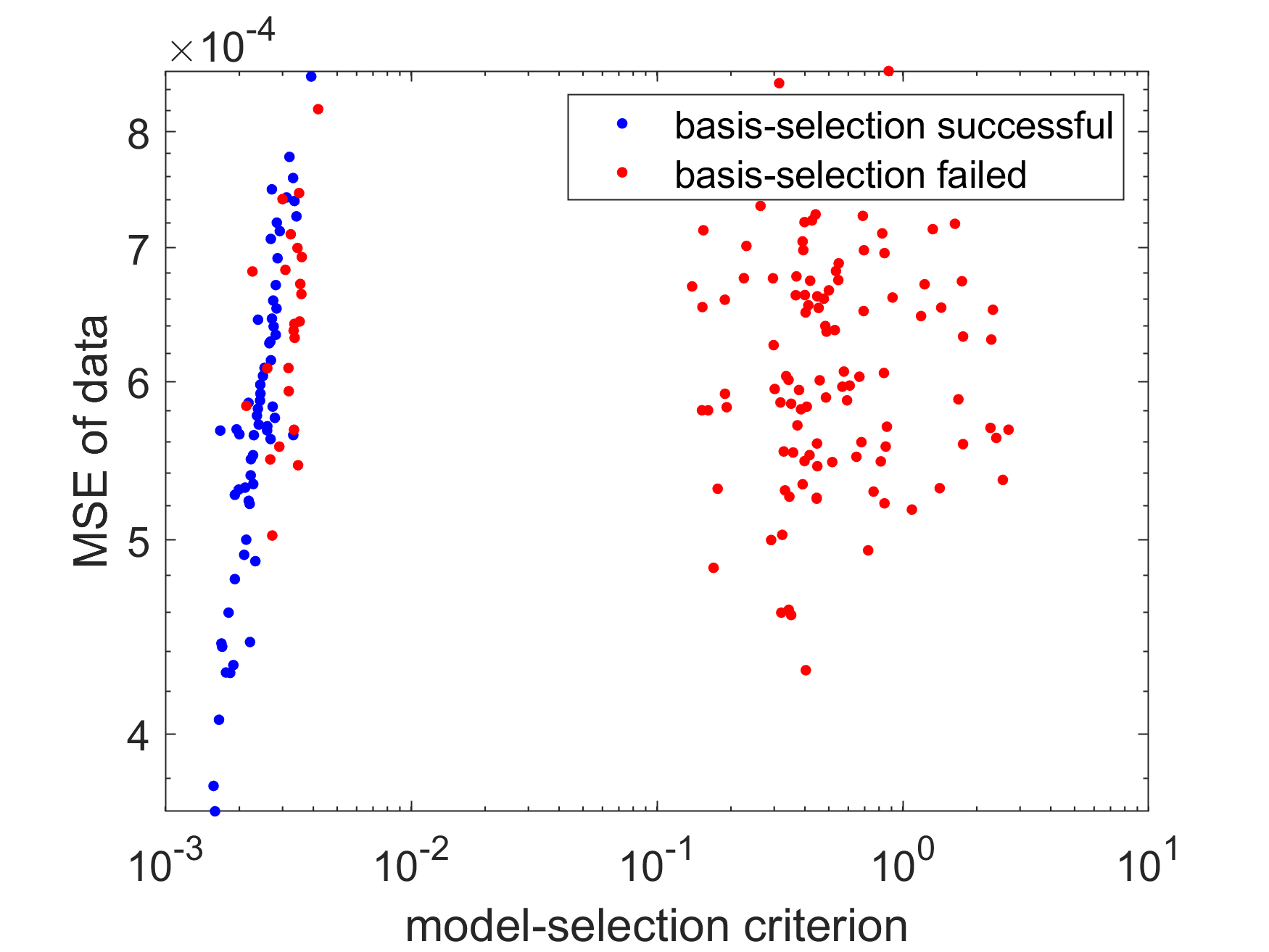}\label{ex18a}}
  (b)\subfloat{\includegraphics[width=7cm]{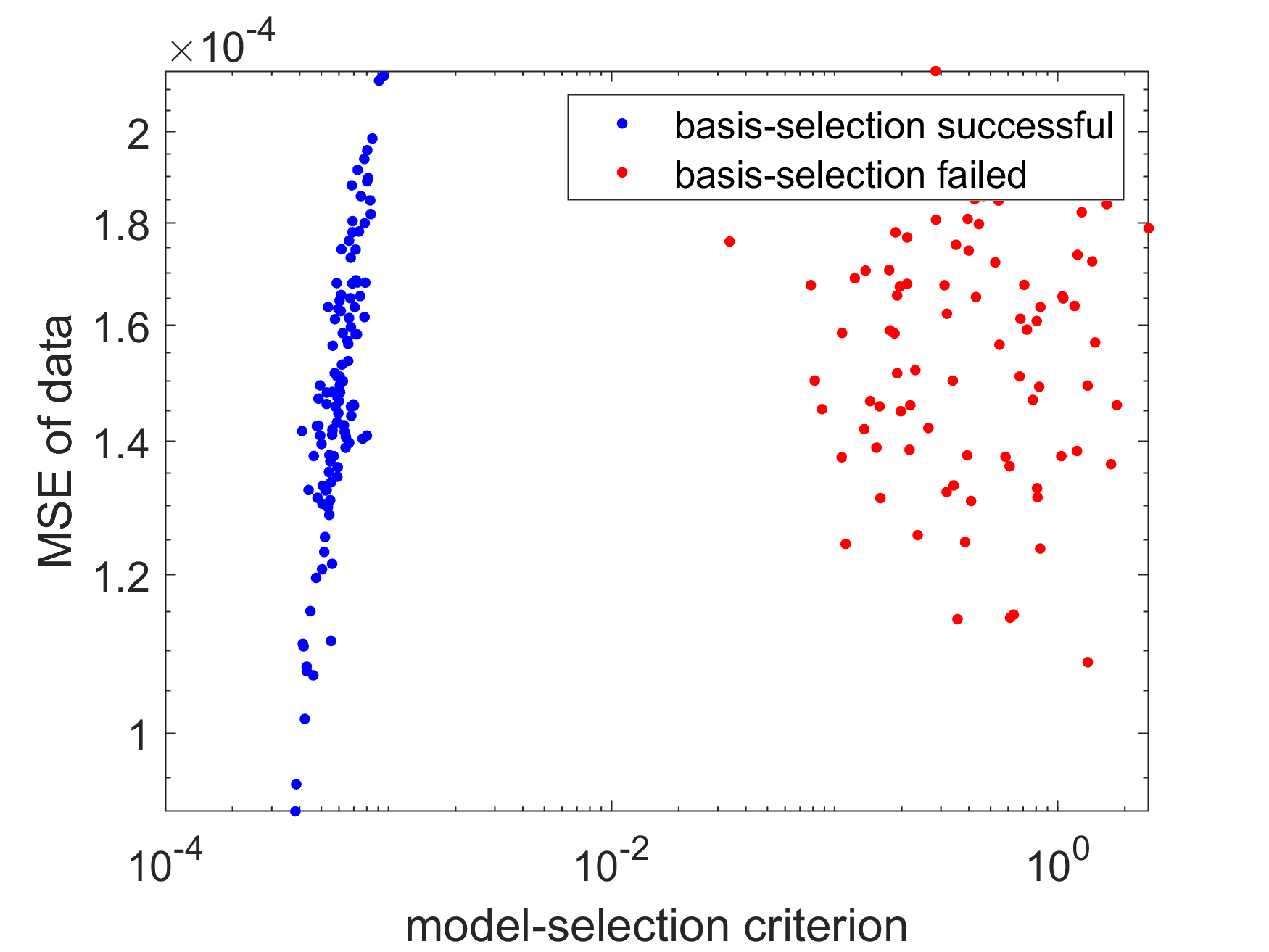}\label{ex18b}}
  \caption{[Predator-prey model] MSE of data vs model-selection criterion. Each point represents the result of a subsample of size $60$. (a) The subsamples are taken from the data in Figure \ref{ex5}. (b) The subsamples are taken from the data in Figure \ref{ex17a} and Figure \ref{ex17b}.}
  \label{ex18}
\end{figure}

Now we visualize how the model-selection criterion is able to select the best subsample. We use the data in Figure \ref{ex5}. The basis-functions are monomials generated by $\{1, x, y\}$ up to degree three ($10$ terms in total). The threshold is set at $0.1$ and the subsampling size is set at $60$. We randomly choose $200$ subsamples and use TSBR (Algorithm \ref{algorithm_tsbr}) on each subsample to discover the equation (\ref{ex3}). For each subsample, we define the mean squared error (MSE) of the data as the mean of $|dx/dt - (1/2) x + (3/2) xy|^2$ within the subsample. TSBR gives a model that either has the same basis-functions as (\ref{ex3}) or not, as well as a model-selection criterion. The result is plotted in Figure \ref{ex18a}. If we use the data in Figure \ref{ex17a} and Figure \ref{ex17b} instead, with the other settings remaining the same, the result would be Figure \ref{ex18b}. In practice, we do not know the correct formula for the model before discovering it, so we cannot calculate the MSE of the data as defined above. However, our model-selection criterion is able to select the best subsample in the sense of the MSE of data and basis-selection successfulness.

\subsection{Example: shallow water equations with outliers}
Consider the following 2-D conservative form of shallow water equations:
\begin{eqnarray}
  \frac{\partial h}{\partial t} + \frac{\partial (hu)}{\partial x} + \frac{\partial (hv)}{\partial y} &=& 0 \label{shallow1} \\
  \frac{\partial (hu)}{\partial t} + \frac{\partial (hu^2 + (1/2) gh^2)}{\partial x} + \frac{\partial (huv)}{\partial y} &=& 0 \label{shallow2} \\
  \frac{\partial (hv)}{\partial t} + \frac{\partial (huv)}{\partial x} + \frac{\partial (hv^2 + (1/2) gh^2)}{\partial y}  &=& 0 \label{shallow3}
\end{eqnarray}
on $(x,y)\in [0,39]\times [0,39]$ and $t\in [0,\infty)$, with reflective boundary condition and a water drop initiating gravity waves, where $h$ is the total fluid column height, $(u,v)$ is the fluid's horizontal flow velocity averaged across the vertical column, and $g = 9.8\,\text{m}\,\text{s}^{-2}$ is the gravitational acceleration. The first equation can be derived from mass conservation, the last two from momentum conservation. Here, we have made the assumption that the fluid density is a constant.

\subsubsection{Data collection}
\begin{figure}[t]\centering
  (a)\subfloat{\includegraphics[width=7cm]{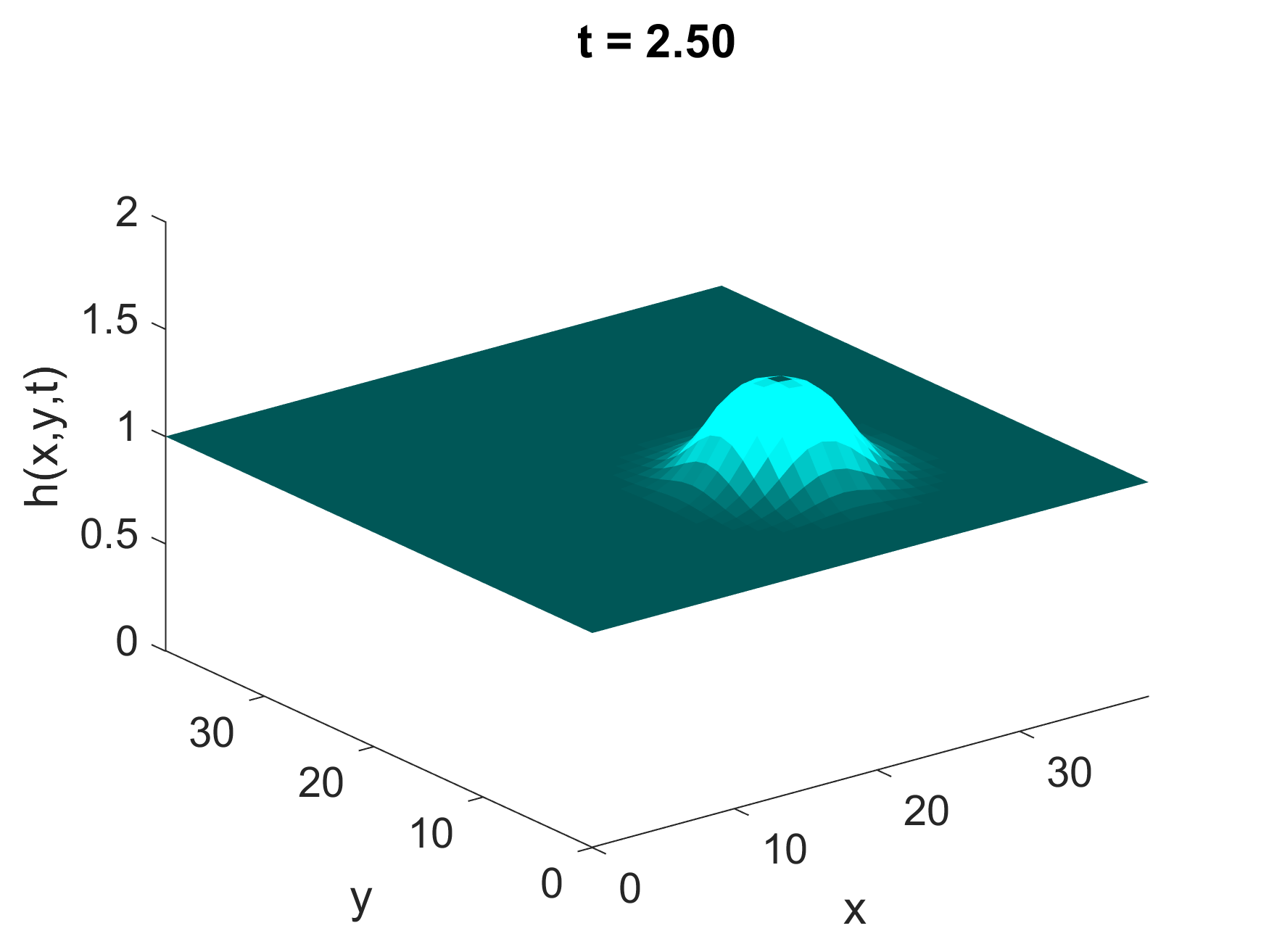}\label{shallow4a}}
  (b)\subfloat{\includegraphics[width=7cm]{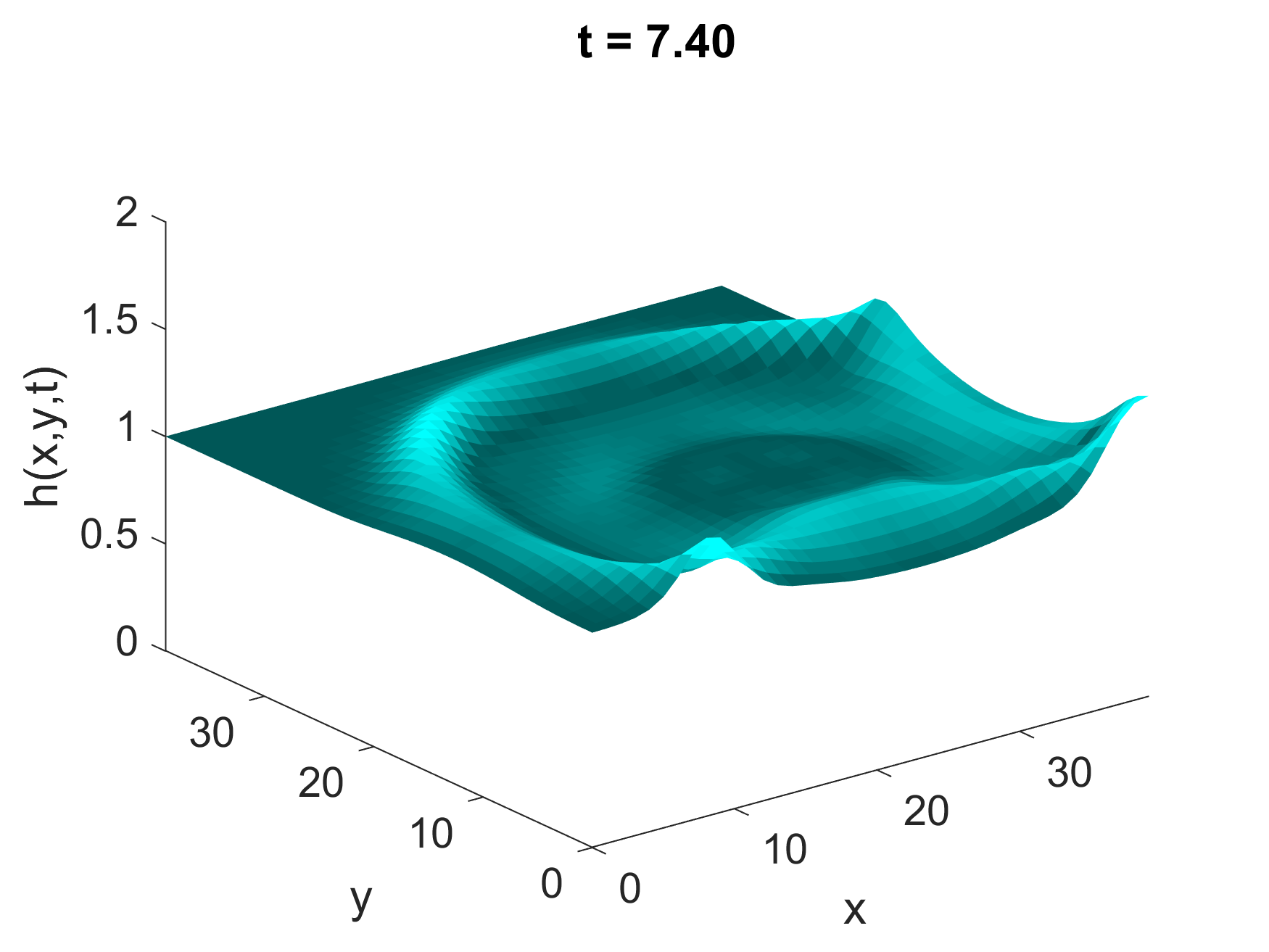}\label{shallow4b}} \\
  (c)\subfloat{\includegraphics[width=7cm]{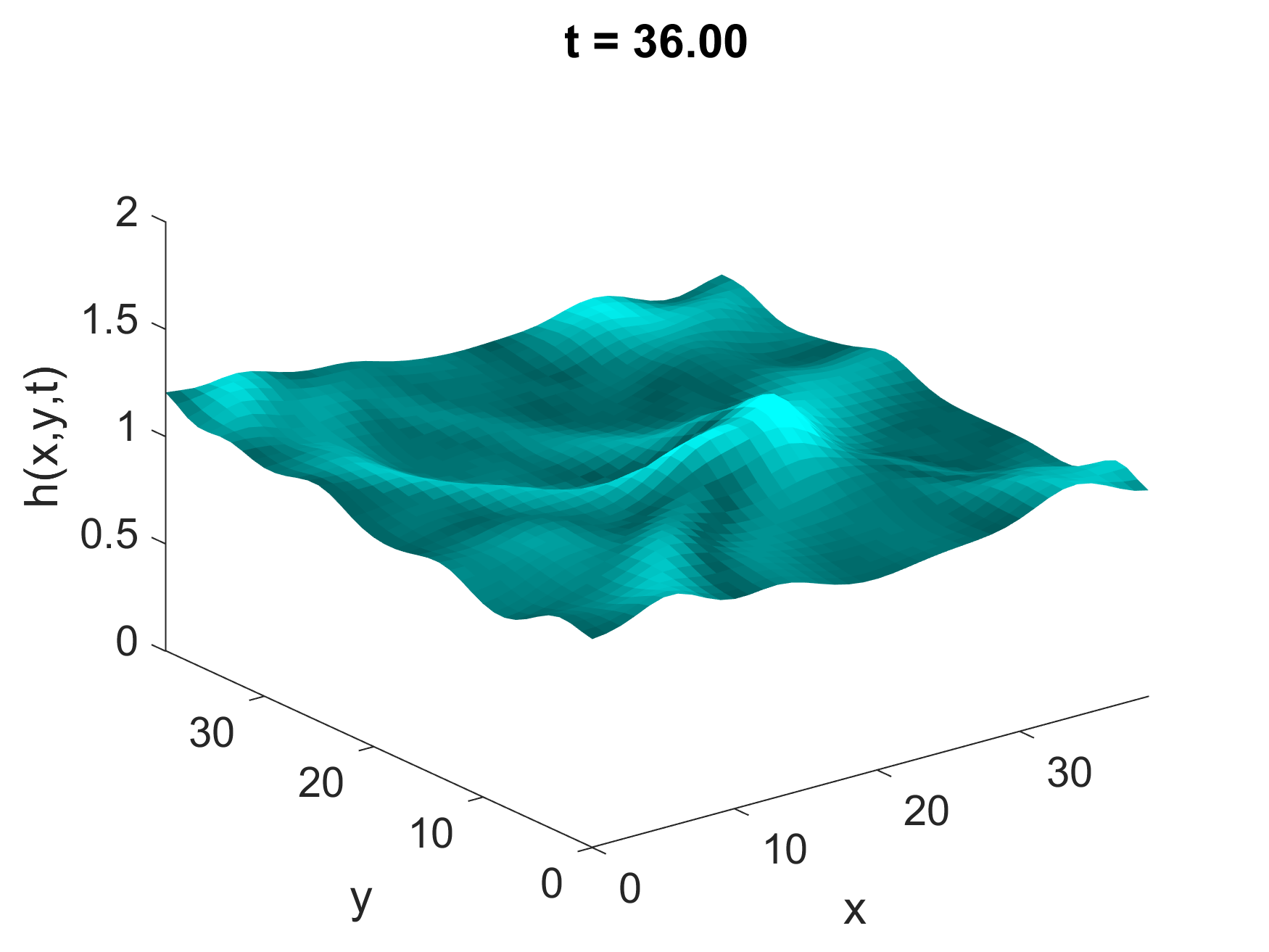}\label{shallow4c}}
  (d)\subfloat{\includegraphics[width=7cm]{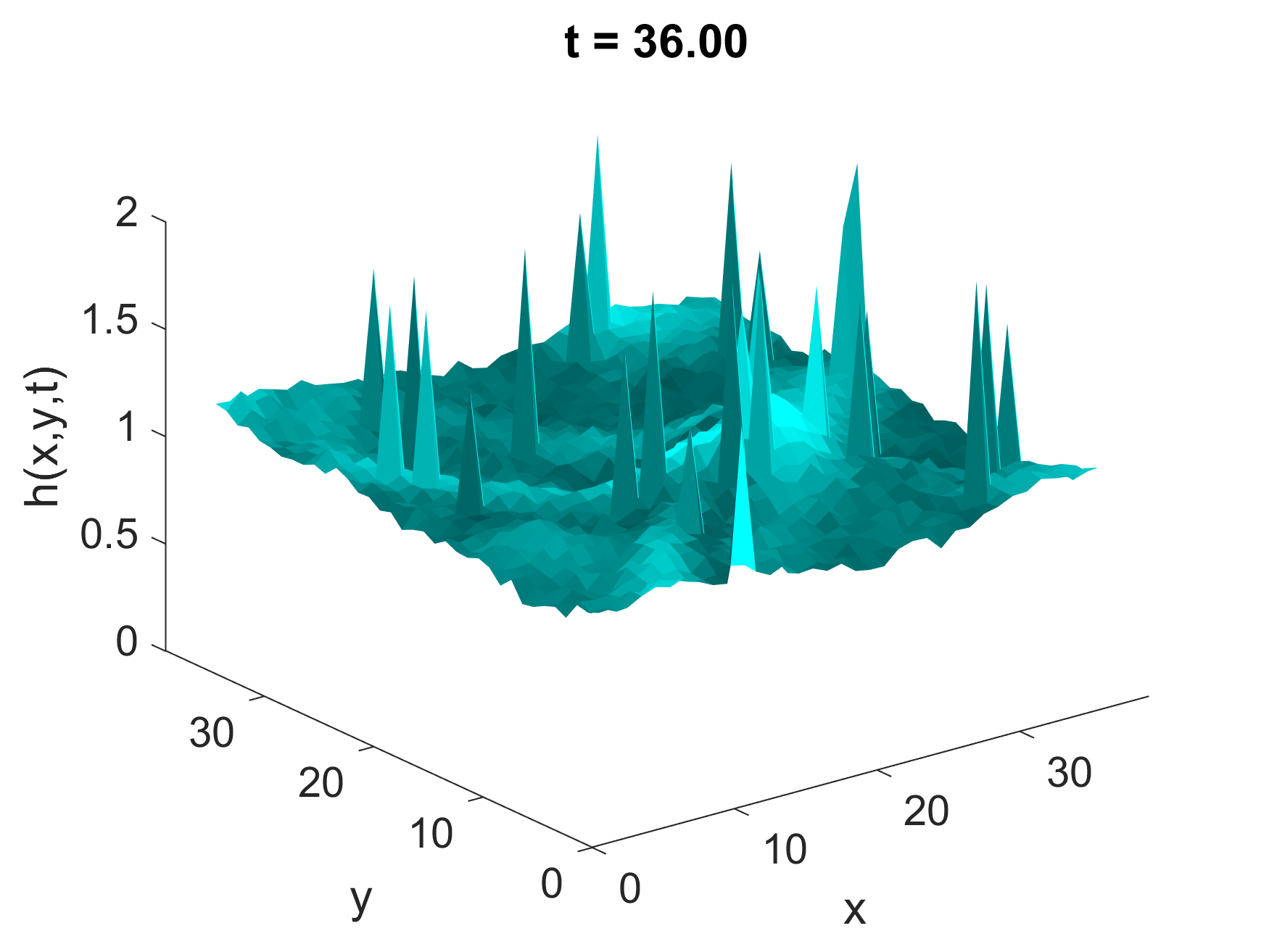}\label{shallow4d}}
  \caption{[Shallow water equations] Surface plot displays height colored by momentum. (a) A water drop falls into the pool. (b) The gravity waves are traveling and being reflected by the boundary. (c) The water surface state when the data are collected. (d) The accessible data are corrupted by noise and outliers.}
  \label{shallow4}
\end{figure}

We generate the numerical solution to the shallow water equations using Lax-Wendroff finite difference method with $\Delta x = \Delta y = 1$ and $\Delta t = 0.02$. See Figure \ref{shallow4}. The data are collected at $t = 36$ and the partial derivatives $\partial h/\partial x$, $\partial u/\partial x$, $\partial v/\partial x$, $\partial h/\partial y$, $\partial u/\partial y$, and $\partial v/\partial y$ are calculated by the three-point central-difference formula. The calculation of the partial derivatives $\partial h/\partial t$, $\partial u/\partial t$, and $\partial v/\partial t$ uses the points from two adjacent time frames. Assume that only the central $36\times 36$ part of the data is made accessible and all the accessible values $h$, $u$, and $v$ are corrupted by independent and identically distributed random noise $\sim \mathcal{N}(0,0.02^2)$. In addition, $2\%$ of the accessible data have the values $h$, $u$, and $v$ corrupted by independent and identically distributed random noise $\sim \mathcal{U}(0.5,1)$ (the uniform distribution on $[0.5,1]$). There are $36\times 36 = 1296$ accessible data points. See Figure \ref{shallow4d}. Thus, the accessible data to discover the model are:
\begin{equation}
    \left\{h_i, u_i, v_i, \left(\frac{\partial h}{\partial t}\right)_i, \left(\frac{\partial u}{\partial t}\right)_i, \left(\frac{\partial v}{\partial t}\right)_i, \left(\frac{\partial h}{\partial x}\right)_i, \left(\frac{\partial u}{\partial x}\right)_i, \left(\frac{\partial v}{\partial x}\right)_i, \left(\frac{\partial h}{\partial y}\right)_i, \left(\frac{\partial u}{\partial y}\right)_i, \left(\frac{\partial v}{\partial y}\right)_i\right\}_{i=1}^{1296},
\end{equation}
$2\%$ of which are outliers.

\subsubsection{Discovery of the model}
We apply the dimensional analysis method introduced in \cite{zhang_robust_2018} to construct the basis-functions of the same dimension as $\partial h/\partial t$ ($\text{m}\,\text{s}^{-1}$) to discover it:
\begin{equation}
\left\{h\frac{\partial u}{\partial x}, h\frac{\partial v}{\partial x}, h\frac{\partial u}{\partial y}, h\frac{\partial v}{\partial y}, u, u\frac{\partial h}{\partial x}, u\frac{\partial h}{\partial y}, v, v\frac{\partial h}{\partial x}, v\frac{\partial h}{\partial y}, \frac{\partial h}{\partial t}\frac{\partial h}{\partial x}, \frac{\partial h}{\partial t}\frac{\partial h}{\partial y}\right\}.
\end{equation}
Similarly, we can construct the basis-functions of the same dimension as $\partial u/\partial t$ and $\partial v/\partial t$ ($\text{m}\,\text{s}^{-2}$) to discover them:
\begin{equation}
  \left\{u\frac{\partial u}{\partial x}, u\frac{\partial v}{\partial x}, u\frac{\partial u}{\partial y}, u\frac{\partial v}{\partial y}, v\frac{\partial u}{\partial x}, v\frac{\partial v}{\partial x}, v\frac{\partial u}{\partial y}, v\frac{\partial v}{\partial y}, \frac{\partial h}{\partial t}\frac{\partial u}{\partial x}, \frac{\partial h}{\partial t}\frac{\partial v}{\partial x}, \frac{\partial h}{\partial t}\frac{\partial u}{\partial y}, \frac{\partial h}{\partial t}\frac{\partial v}{\partial y}, g\frac{\partial h}{\partial x}, g\frac{\partial h}{\partial y}\right\}.
\end{equation}
In this way, the learnt coefficients of each term in the equations are dimensionless. Now we try to discover the shallow water equations (\ref{shallow1}) - (\ref{shallow3}) using threshold sparse Bayesian regression (TSBR, Algorithm \ref{algorithm_tsbr}) and subsampling-based threshold sparse Bayesian regression (SubTSBR, Algorithm \ref{algorithm_subtsbr}). Both TSBR and SubTSBR have the threshold set at $0.1$. In SubTSBR, since there are $1296$ data points and $2\%$ are outliers, if the subsampling size is $100$, then the number of subsamples would be $36$ by (\ref{outlier1}) in order to have confidence level $0.99$. TSBR uses all $1296$ data points at one time to discover the model while SubTSBR does subsampling. Note that the system of equations (\ref{shallow1}) - (\ref{shallow3}) is equivalent to
\begin{eqnarray}
  \frac{\partial h}{\partial t} + h\frac{\partial u}{\partial x} + h\frac{\partial v}{\partial y} + u\frac{\partial h}{\partial x} + v\frac{\partial h}{\partial y} &=& 0 \label{shallow5} \\
  \frac{\partial u}{\partial t} + u\frac{\partial u}{\partial x} + v\frac{\partial u}{\partial y} + g\frac{\partial h}{\partial x} &=& 0 \label{shallow6} \\
  \frac{\partial v}{\partial t} + u\frac{\partial v}{\partial x} + v\frac{\partial v}{\partial y} + g\frac{\partial h}{\partial y} &=& 0. \label{shallow7}
\end{eqnarray}
The numerical results are listed in Table \ref{shallow8}, Table \ref{shallow9}, and Table \ref{shallow10}, from which we can see that SubTSBR has better performance.

\begin{table}[tp]
\centering
\begin{tabular}{|c|r|r|r|r|r|}
\hline
Term                                                         & \multicolumn{1}{c|}{True} & \multicolumn{1}{c|}{TSBR} & \multicolumn{1}{c|}{SubTSBR} \\ \hline
$h\frac{\partial u}{\partial x}$                             & $-1$                      & $-0.974(0.009)$           & $-1.004(0.005)$              \\ \hline
$h\frac{\partial v}{\partial x}$                             &                           &                           &                              \\ \hline
$h\frac{\partial u}{\partial y}$                             &                           &                           &                              \\ \hline
$h\frac{\partial v}{\partial y}$                             & $-1$                      & $-0.958(0.008)$           & $-0.996(0.004)$              \\ \hline
$u$                                                          &                           &                           &                              \\ \hline
$u\frac{\partial h}{\partial x}$                             & $-1$                      & $-0.644(0.075)$           & $-0.929(0.047)$              \\ \hline
$u\frac{\partial h}{\partial y}$                             &                           & $0.489(0.069)$            &                              \\ \hline
$v$                                                          &                           &                           &                              \\ \hline
$v\frac{\partial h}{\partial x}$                             &                           & $0.272(0.073)$            &                              \\ \hline
$v\frac{\partial h}{\partial y}$                             & $-1$                      & $-0.774(0.066)$           & $-0.982(0.043)$              \\ \hline
$\frac{\partial h}{\partial t}\frac{\partial h}{\partial x}$ &                           &                           $-0.240(0.127)$           &                              \\ \hline
$\frac{\partial h}{\partial t}\frac{\partial h}{\partial y}$ &                           & &                              \\ \hline
\end{tabular}
\caption{[Shallow water equations] The true model for $\partial h/\partial t$ (\ref{shallow5}) and the models discovered by threshold sparse Bayesian regression (TSBR) and subsampling-based threshold sparse Bayesian regression (SubTSBR). Every column represents the weights for the terms in the model. Blank means the model does not have the specific term. The numbers inside the parentheses following the weights in TSBR and SubTSBR represent the standard deviation for the weight.} \label{shallow8}
\end{table}

\begin{table}[tp]
\begin{tabular}{|c|r|r|r|r|r|}
\hline
Term                                                         & \multicolumn{1}{c|}{True} & \multicolumn{1}{c|}{TSBR} & \multicolumn{1}{c|}{SubTSBR} \\ \hline
$u\frac{\partial u}{\partial x}$                             & $-1$                      & $-0.621(0.016)$           & $-0.970(0.021)$              \\ \hline
$u\frac{\partial v}{\partial x}$                             &                           & &                              \\ \hline
$u\frac{\partial u}{\partial y}$                             &                           & $0.179(0.019)$            &                              \\ \hline
$u\frac{\partial v}{\partial y}$                             &                           &                           &                              \\ \hline
$v\frac{\partial u}{\partial x}$                             &                           & $0.234(0.017)$            &                              \\ \hline
$v\frac{\partial v}{\partial x}$                             &                           & &                              \\ \hline
$v\frac{\partial u}{\partial y}$                             & $-1$                      & $-0.697(0.019)$           & $-1.010(0.027)$              \\ \hline
$v\frac{\partial v}{\partial y}$                             &                           &                           &                              \\ \hline
$\frac{\partial h}{\partial t}\frac{\partial u}{\partial x}$ &                           &                           &                              \\ \hline
$\frac{\partial h}{\partial t}\frac{\partial v}{\partial x}$ &                           &                           &                              \\ \hline
$\frac{\partial h}{\partial t}\frac{\partial u}{\partial y}$ &                           &                           $-0.141(0.042)$           &                              \\ \hline
$\frac{\partial h}{\partial t}\frac{\partial v}{\partial y}$ &                           &                           &                              \\ \hline
$g\frac{\partial h}{\partial x}$                             & $-1$                      & $-1.010(0.001)$           & $-0.999(0.001)$              \\ \hline
$g\frac{\partial h}{\partial y}$                             &                           &                           &                              \\ \hline
\end{tabular}
\caption{[Shallow water equations] The true model for $\partial u/\partial t$ (\ref{shallow6}) and the discovered models. All other interpretations are the same as Table \ref{shallow8}.} \label{shallow9}
\end{table}

\begin{table}[tp]
\begin{tabular}{|c|r|r|r|r|r|}
\hline
Term                                                         & \multicolumn{1}{c|}{True} & \multicolumn{1}{c|}{TSBR} & \multicolumn{1}{c|}{SubTSBR} \\ \hline
$u\frac{\partial u}{\partial x}$                             &                           &                           &                              \\ \hline
$u\frac{\partial v}{\partial x}$                             & $-1$                      & $-1.271(0.267)$           & $-1.008(0.022)$              \\ \hline
$u\frac{\partial u}{\partial y}$                             &                           &                           $0.524(0.266)$            &                              \\ \hline
$u\frac{\partial v}{\partial y}$                             &                           & $0.283(0.012)$            &                              \\ \hline
$v\frac{\partial u}{\partial x}$                             &                           &                           &                              \\ \hline
$v\frac{\partial v}{\partial x}$                             &                           & $0.160(0.017)$            &                              \\ \hline
$v\frac{\partial u}{\partial y}$                             &                           & &                              \\ \hline
$v\frac{\partial v}{\partial y}$                             & $-1$                      & $-0.644(0.014)$           & $-0.962(0.017)$              \\ \hline
$\frac{\partial h}{\partial t}\frac{\partial u}{\partial x}$ &                           & &                              \\ \hline
$\frac{\partial h}{\partial t}\frac{\partial v}{\partial x}$ &                           & &                              \\ \hline
$\frac{\partial h}{\partial t}\frac{\partial u}{\partial y}$ &                           &                           &                              \\ \hline
$\frac{\partial h}{\partial t}\frac{\partial v}{\partial y}$ &                           & &                              \\ \hline
$g\frac{\partial h}{\partial x}$                             &                           &                           &                              \\ \hline
$g\frac{\partial h}{\partial y}$                             & $-1$                      & $-1.009(0.001)$           & $-1.001(0.001)$              \\ \hline
\end{tabular}
\caption{[Shallow water equations] The true model for $\partial v/\partial t$ (\ref{shallow7}) and the discovered models. All other interpretations are the same as Table \ref{shallow8}.} \label{shallow10}
\end{table}

\section{Merits and applications of discovering differential equations from data}
\label{app}
In this section, we demonstrate some merits of discovering differential equations from data:
\begin{enumerate}[label=(\alph*)]
  \item Integration of the data from different experiments into meaningful and valuable information. In many cases, collecting enough data from a single experiment is difficult to achieve due to limited resources. For instance, the experiments may need to be done at multiple different times and/or different locations. When the data are from multiple experiments, although they are generated by the same model, the initial condition and/or boundary condition used to generate them may be different or even unmeasurable. This is a challenge for traditional interpolation and regression methods. A significant advantage of the method of discovering governing differential equations is that the data are allowed to be from different experiments, as long as the model governing them is the same. On top of that, if the initial condition and boundary condition can be formulated into algebraic equations and we are given data at the initial state and boundary, we may symbolize the initial condition and boundary condition into the form (\ref{intro4}) and discover them using the subsampling-based threshold sparse Bayesian regression algorithm. The only difference in this case is that the algebraic equations do not have any derivative term. Finally, with the discovered differential equation, initial condition, and boundary condition, we may reconstruct the solutions to the model and make predictions.
  \item Prediction of the generalized dynamics in broader areas when the data are collected within a restricted domain. The solution to the dynamics may have different behavior in different areas. If the data are collected within a restricted domain, traditional linear regression may not be able to capture the behavior or make predictions outside that domain. This becomes even more challenging when the dynamics have bifurcations. The method of discovering differential equations is able to predict the generalized dynamics, because the same differential equations that govern the dynamics inside the restricted domain will govern the outside as well. Experiments in labs sometimes have to be done within a certain range of condition, such as initial condition, but real-world applications are in different scales. Data analysis by discovering differential equations can be helpful since it is able to generalize to broader range of condition.
  \item Extrapolation. When the model is correctly discovered, it performs well in extrapolation as we might expect.
\end{enumerate}

We demonstrate these merits in the two following examples by comparing the results from discovering differential equations with the results from traditional linear regression without discovering differential equations. We study these examples in the presence of noise and/or outliers.

\subsection{Example: heat diffusion with random initial and boundary condition}
Consider the following 1-D heat diffusion equation:
\begin{equation} \label{heat1}
    \frac{\partial u}{\partial t} = \frac{1}{2} \frac{\partial^2 u}{\partial x^2}
\end{equation}
on $x\in [0,5]$ and $t\in [0,\infty)$ with random initial condition:
\begin{equation} \label{heat2}
    u(x,0) = -\frac{1}{2} \xi_1 x (x - 5)
\end{equation}
and random boundary condition:
\begin{eqnarray}
    u(0,t) &=& \xi_2 \sin{(2t)} - \xi_3^2 \cos{t} + \xi_3^2 \label{heat3} \\
    u(5,t) &=& \xi_2 \xi_3 \sin{t} - \xi_3 \sin{(t + \frac{\pi}{4})} + \frac{\xi_3\sqrt{2}}{2}, \label{heat4}
\end{eqnarray}
where $\xi_1, \xi_2, \xi_3$ are independent random variables:
\begin{itemize}
    \item $\xi_1 \sim \mathcal{U}(0,1)$, the uniform distribution on $[0,1]$;
    \item $\xi_2 \sim \mathcal{U}(0,1)$, the uniform distribution on $[0,1]$;
    \item $\xi_3 \sim \mathcal{N}(0,0.5^2)$, the normal distribution with mean $0$ and standard deviation $0.5$.
\end{itemize}
When $\xi_1 = \xi_2 = \xi_3 = 0.5$, the solution to the heat diffusion equation is plotted in Figure \ref{heat10a}.

\subsubsection{Data collection and discovery of the model}
\begin{figure}[p]
  \centering
  (a)\subfloat{\includegraphics[width=15cm]{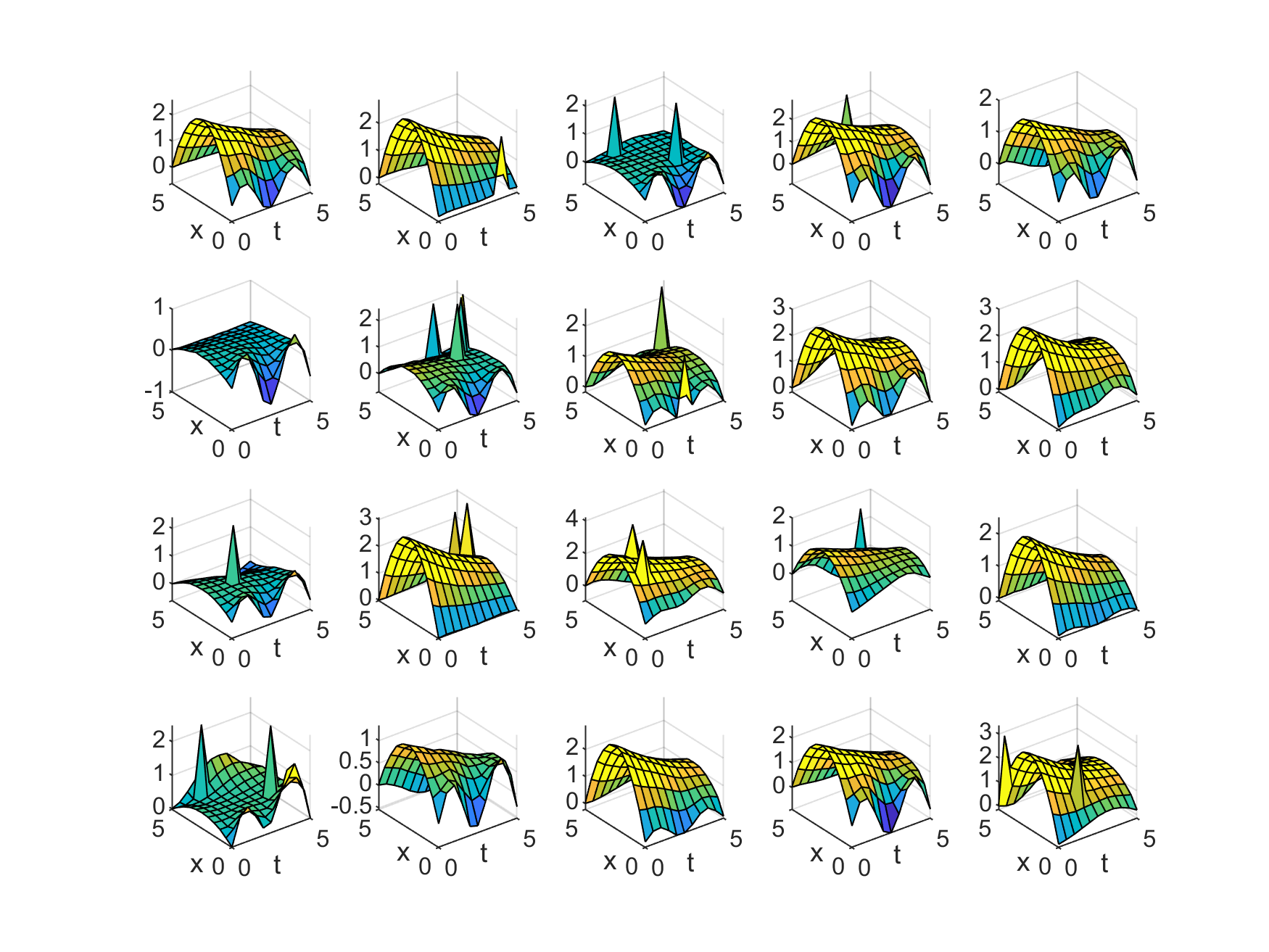}\label{heat6a}} \\
  (b)\subfloat{\includegraphics[width=7cm]{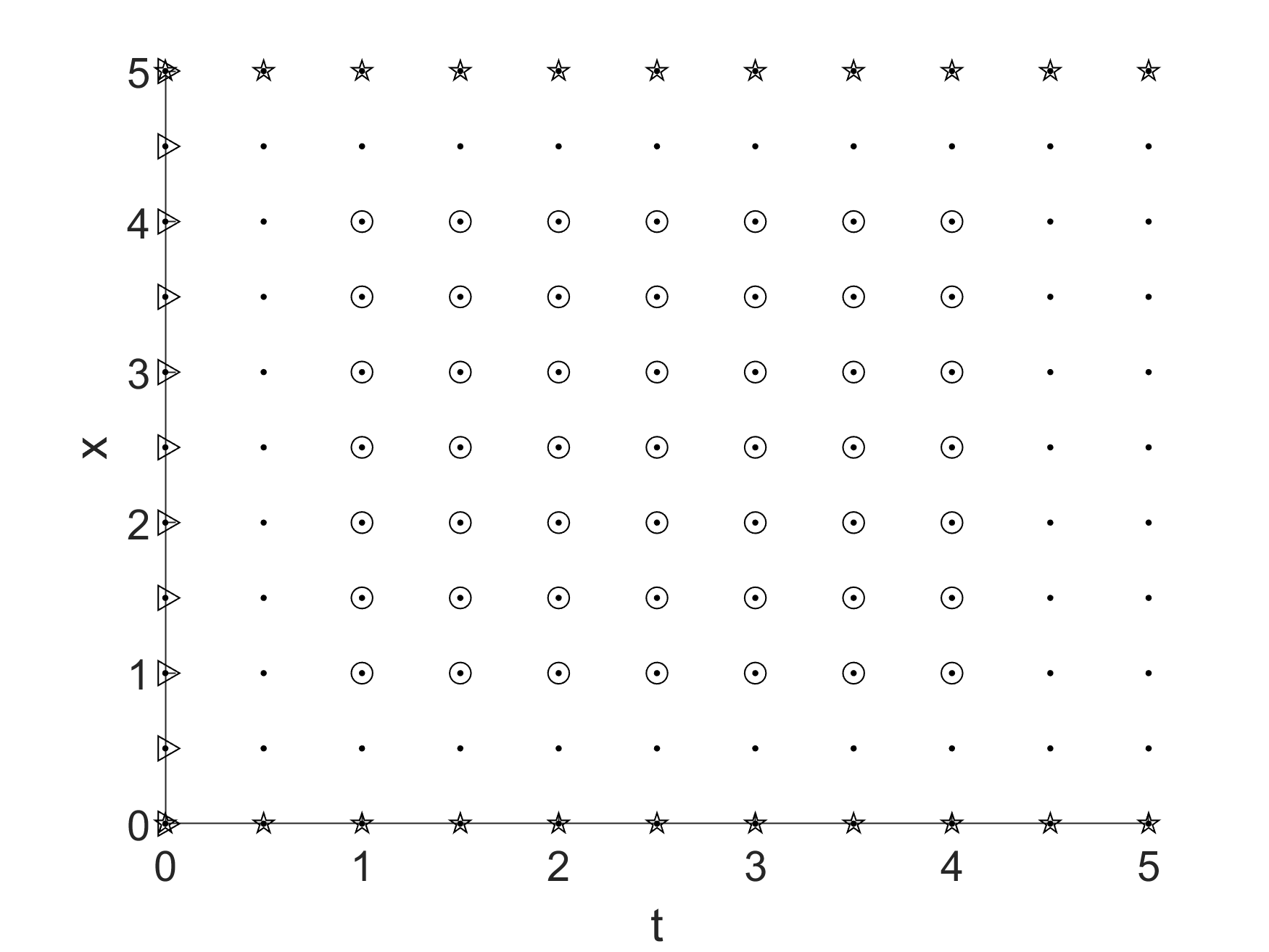} \label{heat6b}}
  \caption{[Heat diffusion] (a) The data are generated by (\ref{heat1}) - (\ref{heat4}) with $20$ sets of independent random variables $\{\xi_1, \xi_2, \xi_3\}$ and corrupted by noise and outliers. (b) Grid points: [$\bullet$]: all data points, [$\bigcirc$]: data points used to discover PDE, [$\star$]: data points used to discover boundary condition, [$\rhd$]: data points used to discover initial condition.} \label{heat6}
\end{figure}

We generate $20$ solutions by (\ref{heat1}) - (\ref{heat4}) with $20$ sets of independent random variables $\{\xi_1, \xi_2, \xi_3\}$. Then we collect the data at the grid points in the domain $x\in [0,5]$ and $t\in [0,5]$ illustrated in Figure \ref{heat6b}. There are $11\times 11\times 20 = 2420$ data points. Next, we calculate derivatives using the five-point central-difference formula. Note that the derivatives are only calculated at the interior grid points (marked as [$\bigcirc$] in Figure \ref{heat6b}). Then all the data are corrupted by independent and identically distributed random noise $\sim \mathcal{N}(0,0.01^2)$ and some of them are further corrupted and become outliers. See Figure \ref{heat6a}. Next, we discover the PDE using our algorithm SubTSBR with subsampling size $245$, which is one fourth of all interior grid points, and $300$ subsamples. The basis-functions are monomials generated by $\{1, x, t, u, \partial u/\partial x, \partial^2 u/\partial x^2\}$ up to degree $3$. There are $56$ terms. The result is:
\begin{equation} \label{heat5}
    \frac{\partial u}{\partial t} = 0.498\,\frac{\partial^2 u}{\partial x^2}.
\end{equation}

After that, we discover the boundary condition using SubTSBR with subsampling size $55$, which is one fourth of all lower boundary points or upper boundary points, and $300$ subsamples. The basis-functions are monomials generated by $\{1, \xi_2, \xi_3, t, \sin{t}, \cos{t}\}$ up to degree $3$. There are $56$ terms. The result is:
\begin{eqnarray}
    u(0,t) &=& 1.006\,\xi_3^2 +  2.009\,\xi_2 \sin{t}\cos{t} - 1.010\,\xi_3^2 \cos{t} \label{heat7} \\
    u(5,t) &=& 0.714\,\xi_3 - 0.693\,\xi_3 \sin{t} - 0.706\,\xi_3 \cos{t} + 0.993\,\xi_2 \xi_3 \sin{t}. \label{heat8}
\end{eqnarray}

Next, we discover the initial condition using SubTSBR with subsampling size $55$, which is one fourth of all initial points, and $300$ subsamples. The basis-functions are monomials generated by $\{1, \xi_1, x, \sin{x}, \cos{x}\}$ up to degree $3$. There are $35$ terms. The result is:
\begin{equation} \label{heat9}
    u(x,0) = 2.494\,\xi_1 x - 0.499\,\xi_1 x^2.
\end{equation}

\subsubsection{Prediction}
\begin{figure}[p]\centering
  (a)\subfloat{\includegraphics[width=12cm]{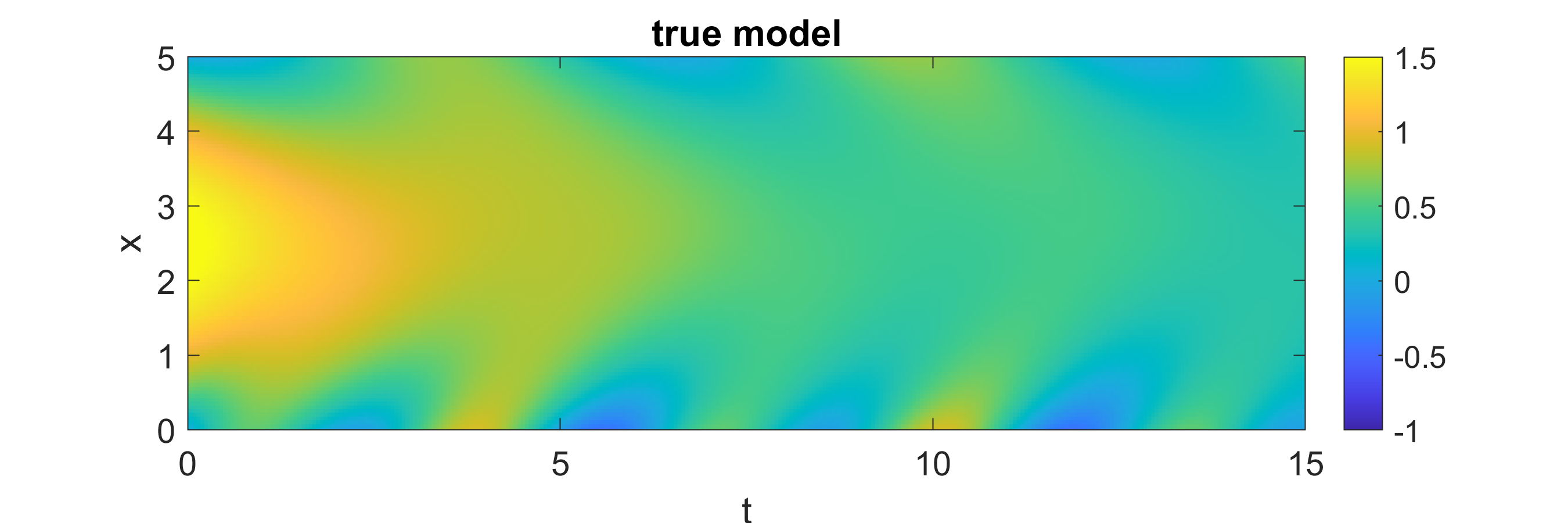}\label{heat10a}} \\
  (b)\subfloat{\includegraphics[width=12cm]{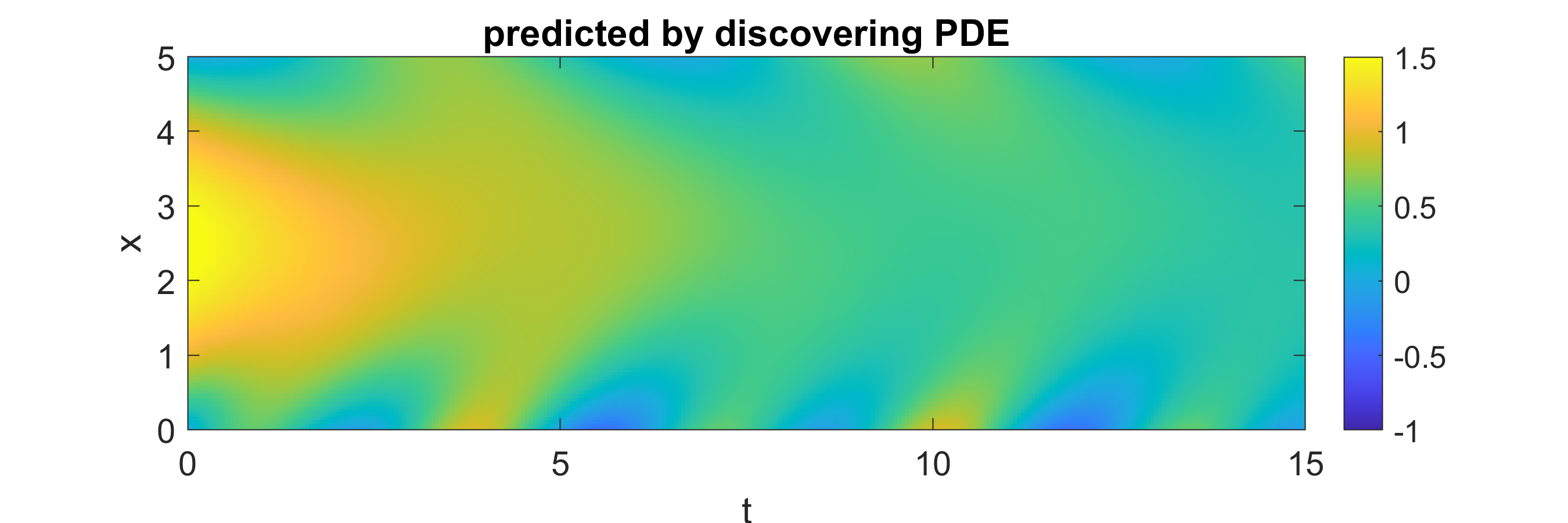}\label{heat10b}} \\
  (c)\subfloat{\includegraphics[width=12cm]{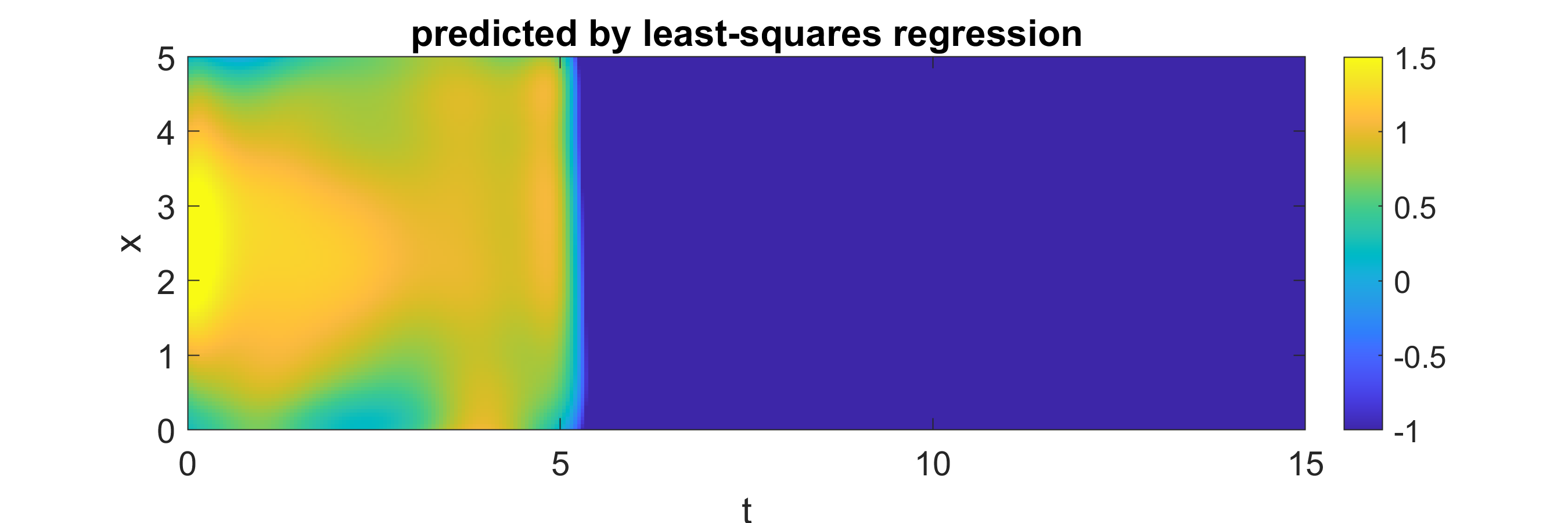}\label{heat10c}}
  \caption{[Heat diffusion] (a) The true model solved by (\ref{heat1}) with initial and boundary condition (\ref{heat2}) - (\ref{heat4}). (b) The solution predicted by discovering PDE, solved by (\ref{heat5}) with initial and boundary condition (\ref{heat7}) - (\ref{heat9}). (c) The solution predicted by least-squares regression. All with $\xi_1 = \xi_2 = \xi_3 = 0.5$.}
  \label{heat10}
\end{figure}

\begin{figure}[p]\centering
  (a)\subfloat{\includegraphics[width=7cm]{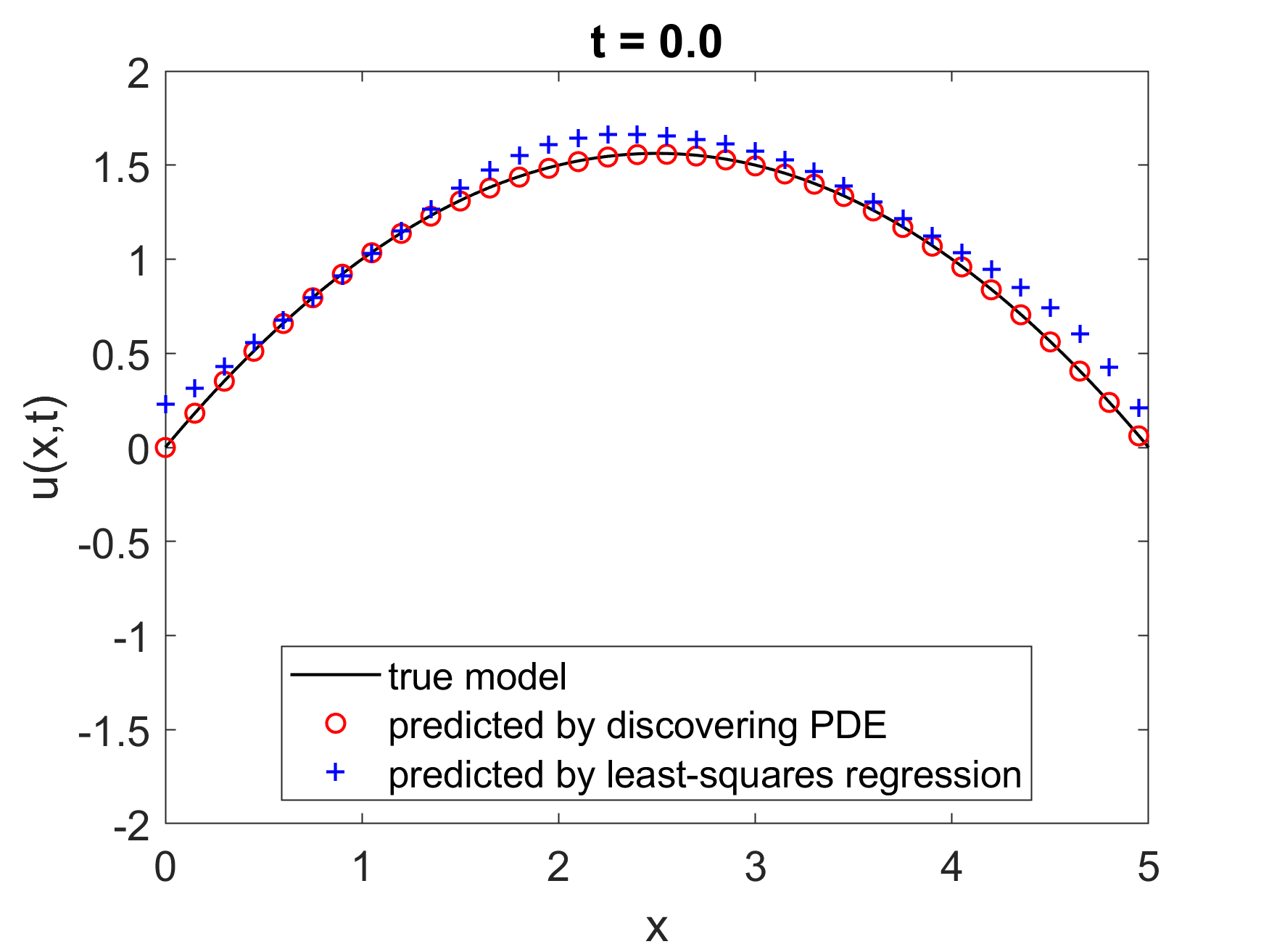}}
  (b)\subfloat{\includegraphics[width=7cm]{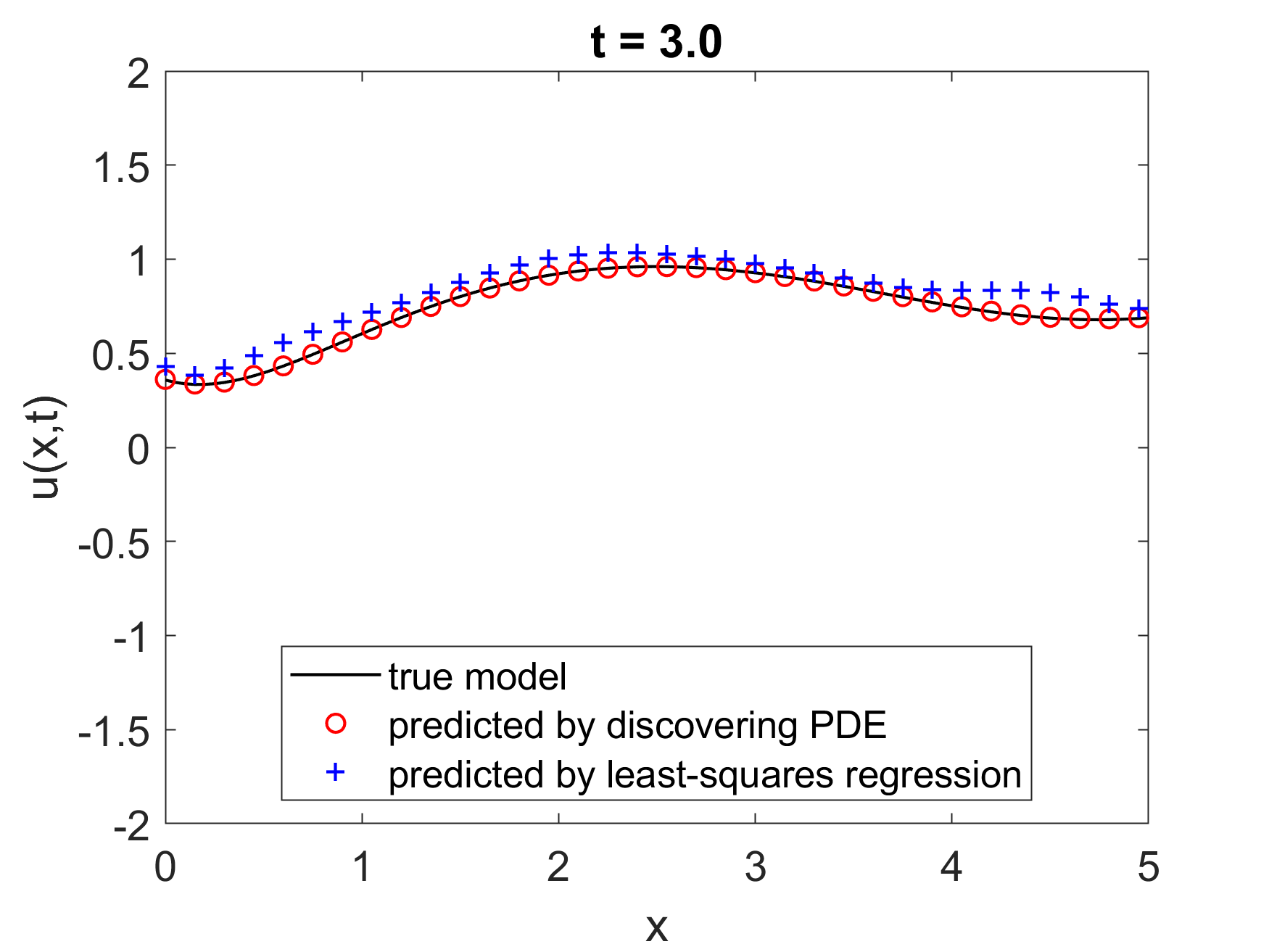}} \\
  (c)\subfloat{\includegraphics[width=7cm]{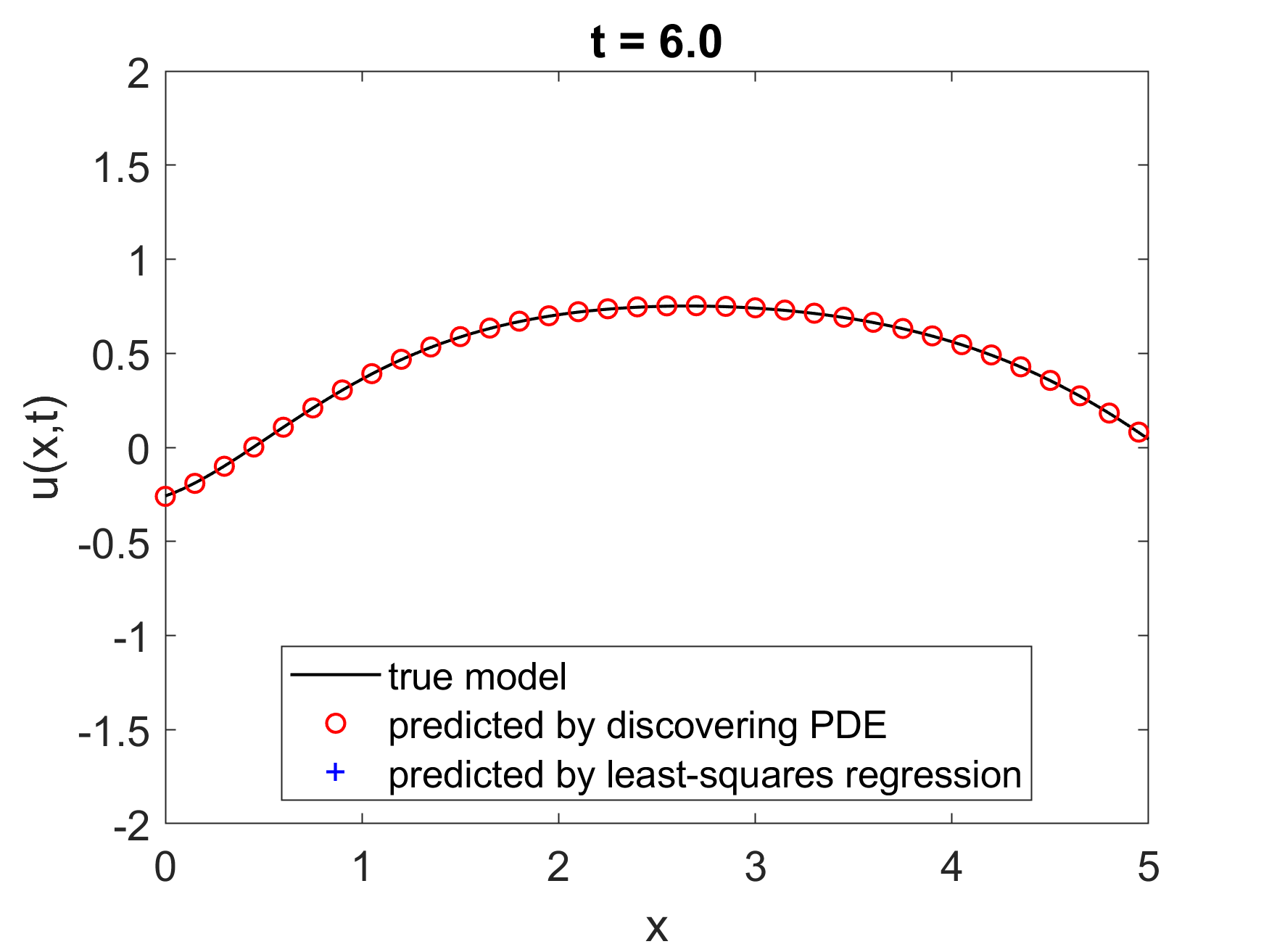}}
  (d)\subfloat{\includegraphics[width=7cm]{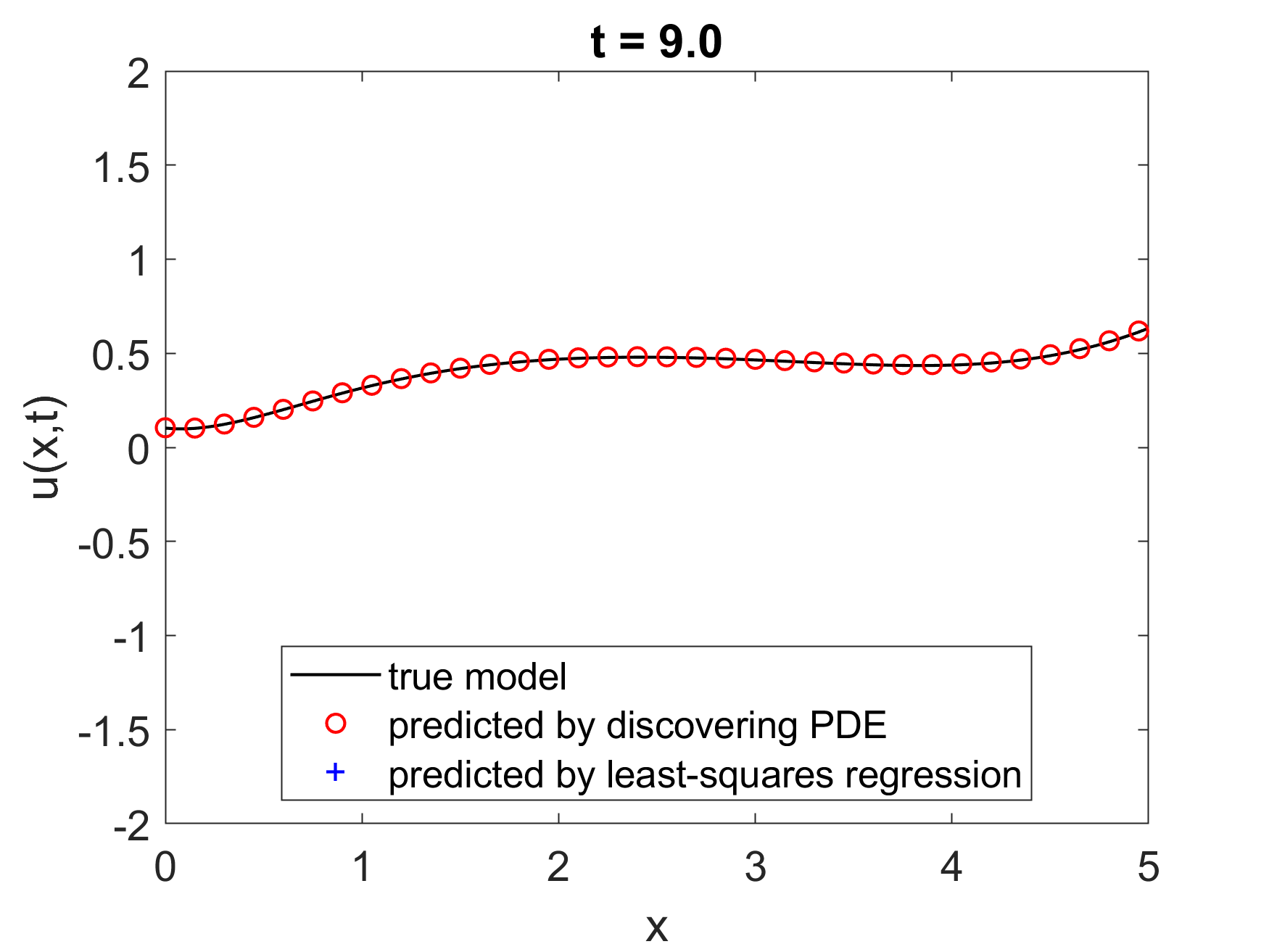}} \\
  (e)\subfloat{\includegraphics[width=7cm]{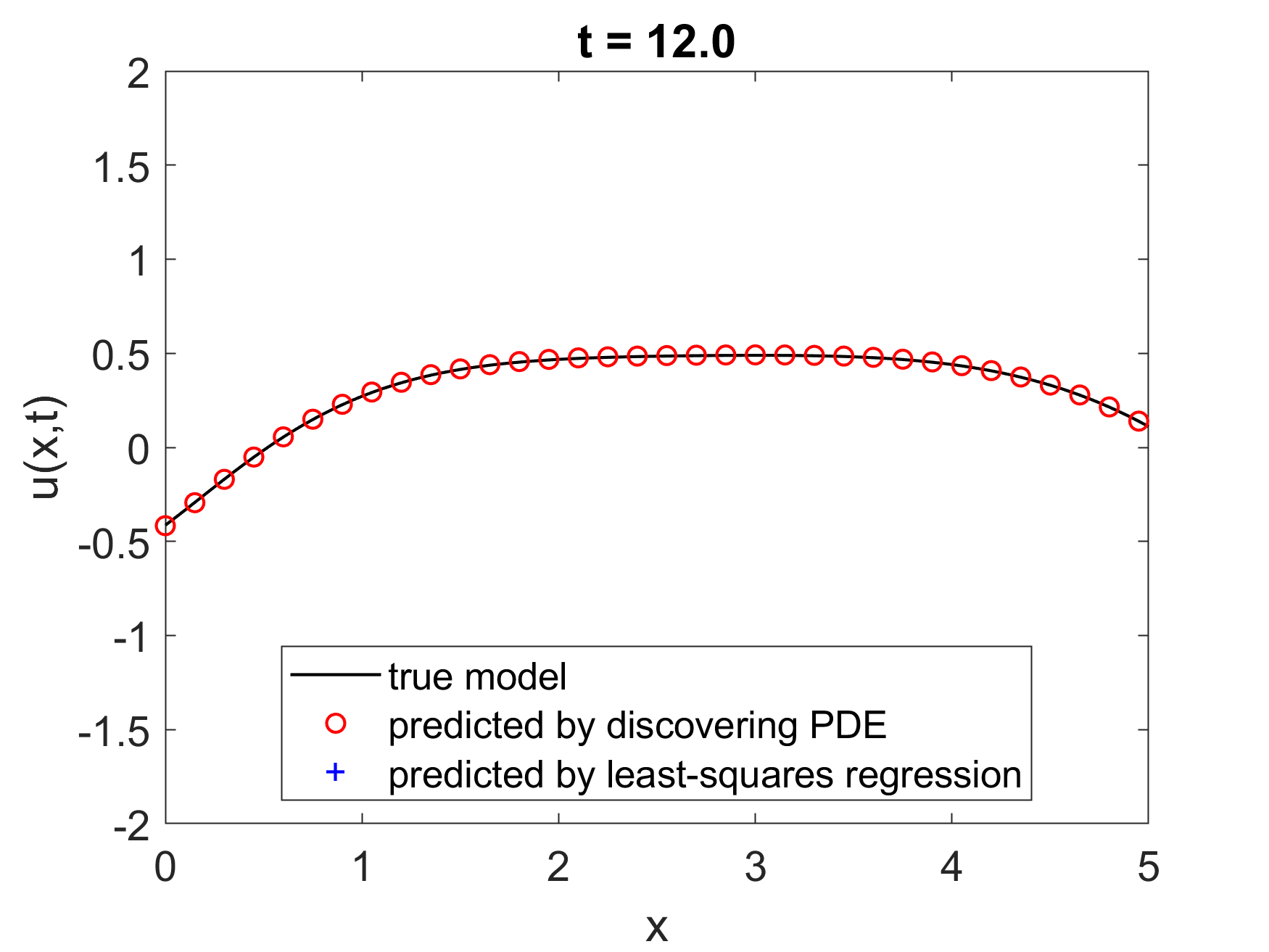}}
  (f)\subfloat{\includegraphics[width=7cm]{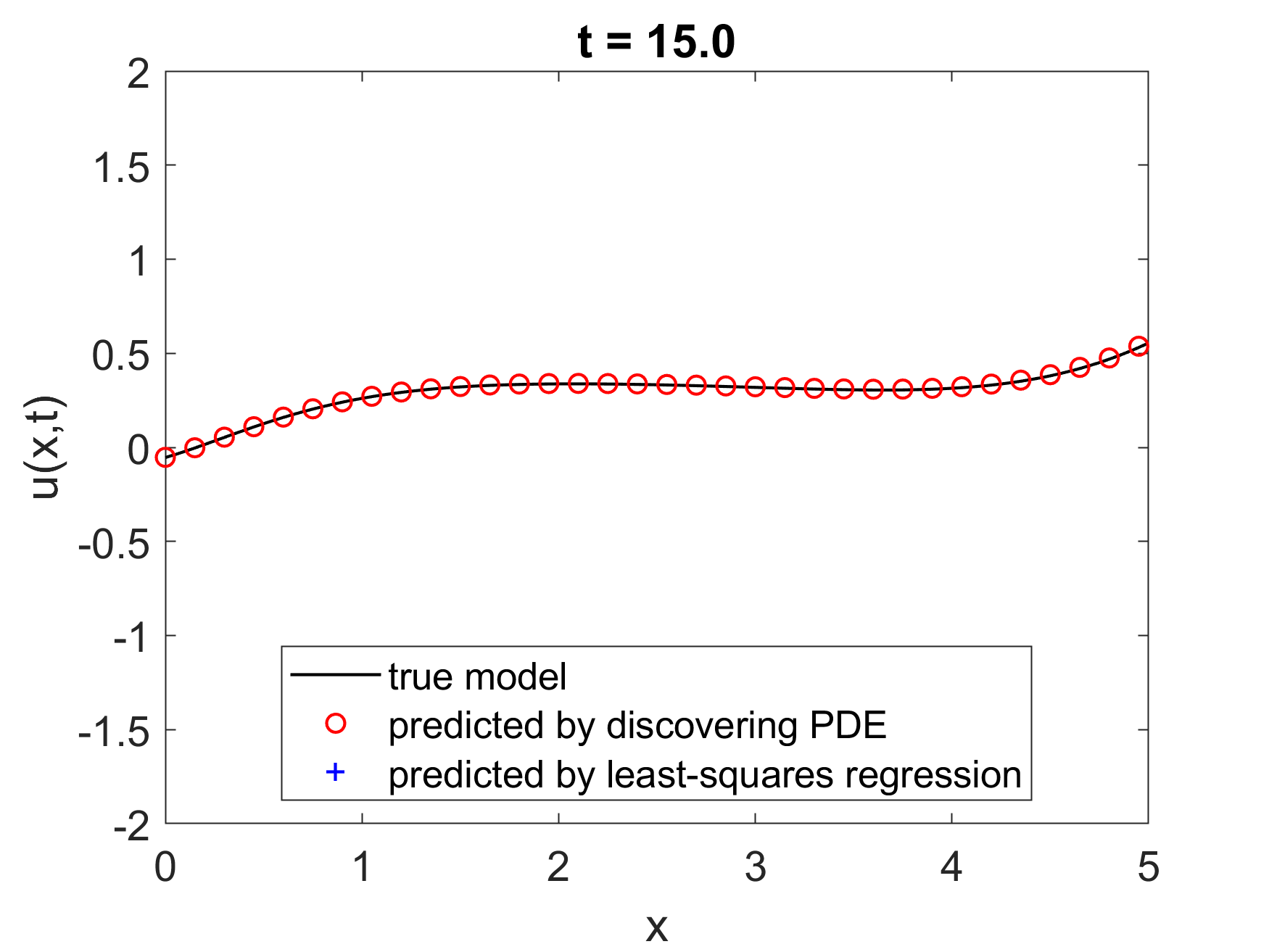}}
  \caption{[Heat diffusion] The true model, the solution predicted by discovering PDE, and the solution predicted by least-squares regression, at different time $t$. All settings are the same as Figure \ref{heat10}. In (c)(d)(e)(f), the predictions by least-squares regression are outside of the axes limits.}
  \label{heat11}
\end{figure}

Now we predict the solution when $\xi_1 = \xi_2 = \xi_3 = 0.5$. Fix the $\xi_1, \xi_2, \xi_3$ values in (\ref{heat7}) - (\ref{heat9}) and solve the PDE (\ref{heat5}) on $x\in [0,5]$ and $t\in [0,15]$. We get the solution to the heat diffusion equation predicted by discovering PDE. See Figure \ref{heat10b}. The true model is solved by (\ref{heat1}) with $\xi_1 = \xi_2 = \xi_3 = 0.5$ in (\ref{heat2}) - (\ref{heat4}). Its solution is displayed in Figure \ref{heat10a}. As a comparison, the solution predicted by least-squares regression is displayed in Figure \ref{heat10c}, where $u(x,t,\xi_1,\xi_2,\xi_3)$ is fitted by a linear combination of monomials generated by $\{1, x, \sin{x}, \cos{x}, t, \sin{t}, \cos{t}, \xi_1, \xi_2, \xi_3\}$ up to degree $3$. There are $220$ terms. Here we use the full data set for discovering PDE to do the regression. The solution is drawn with fixed $\xi_1 = \xi_2 = \xi_3 = 0.5$: $u(x,t,0.5,0.5,0.5)$. Figure \ref{heat10} shows that both methods of discovering PDE and least-squares regression approximates the true model well on $t\in [0,5]$, but discovering PDE predicts the true model much better on $t\in (5,15]$. Least-squares regression starts to fail when $t > 5$ because the data are collected within $t\in [0,5]$ and least-squares regression is an interpolation method. It does not work well outside the region with known data. In contrast, discovering PDE is an extrapolation method, which works beyond the original observation range. See Figure \ref{heat11} for the solution at different time $t$ and Table \ref{heat12} for the mean squared error (MSE) at different time $t$.

\begin{table}[p]
\begin{tabular}{|c|c|c|}
  \hline
  $t$ & MSE by discovering PDE & MSE by least-squares regression \\ \hline
  $0$ & $0.000009$ & $0.010271$ \\ \hline
  $3.0$ & $0.000004$ & $0.007247$ \\ \hline
  $6.0$ & $0.000002$ & $975.998958$ \\ \hline
  $9.0$ & $0.000012$ & $1952956.268589$ \\ \hline
  $12.0$ & $0.000004$ & $5552302.080802$ \\ \hline
  $15.0$ & $0.000013$ & $28323673.179639$ \\ \hline
\end{tabular}
\caption{[Heat diffusion] Mean squared error (MSE) of the predicted solutions in Figure \ref{heat10} by discovering PDE and least-squares regression at different time $t$.} \label{heat12}
\end{table}

Note that we did not use any data generated from $\xi_1 = \xi_2 = \xi_3 = 0.5$ to discover the PDE, initial condition, or boundary condition. Instead, we used the data generated from $20$ random sets of $\{\xi_1, \xi_2, \xi_3\}$. Predictions at other $\xi_1, \xi_2, \xi_3$ values can be derived in the same way by fixing the $\xi_1, \xi_2, \xi_3$ values in (\ref{heat7}) - (\ref{heat9}) and solving (\ref{heat5}). The prediction at $\{\xi_1, \xi_2, \xi_3\}$ does not need any data generated from the same $\{\xi_1, \xi_2, \xi_3\}$.

Also note that we do not need any information about $\xi_1, \xi_2, \xi_3$ to discover the PDE. If the values of $\xi_1, \xi_2, \xi_3$ are unknown, we can still discover the PDE but are unable to discover the initial condition or boundary condition in this example. If given a new initial condition and boundary condition, we can still make prediction using the discovered PDE, while we are not able to do so using regression methods.

\subsection{Example: fish-harvesting problem with bifurcations}
\begin{figure}[t]\centering
  (a)\subfloat{\includegraphics[width=7cm]{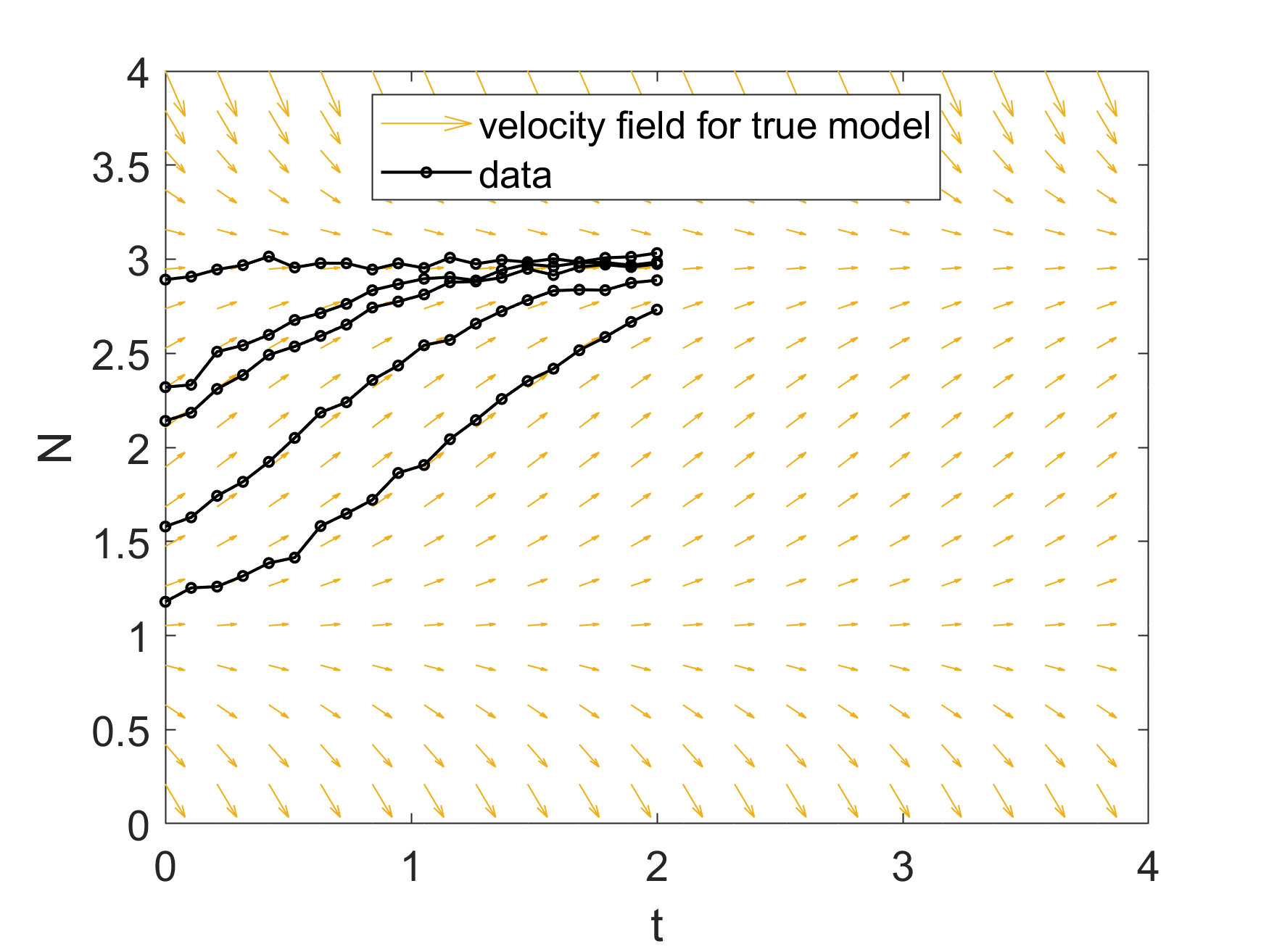}\label{fish2a}}
  (b)\subfloat{\includegraphics[width=7cm]{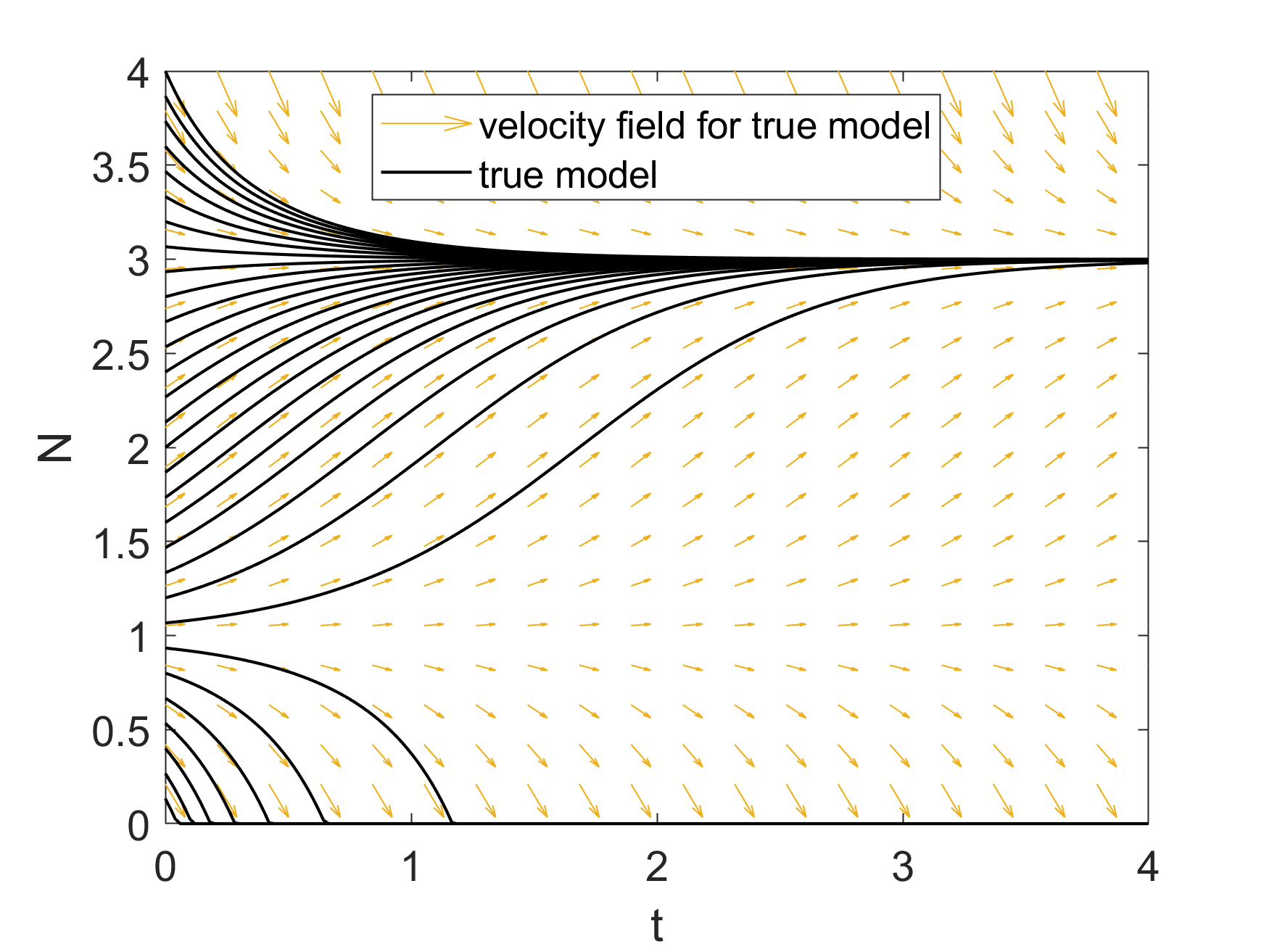}\label{fish2b}} \\
  (c)\subfloat{\includegraphics[width=7cm]{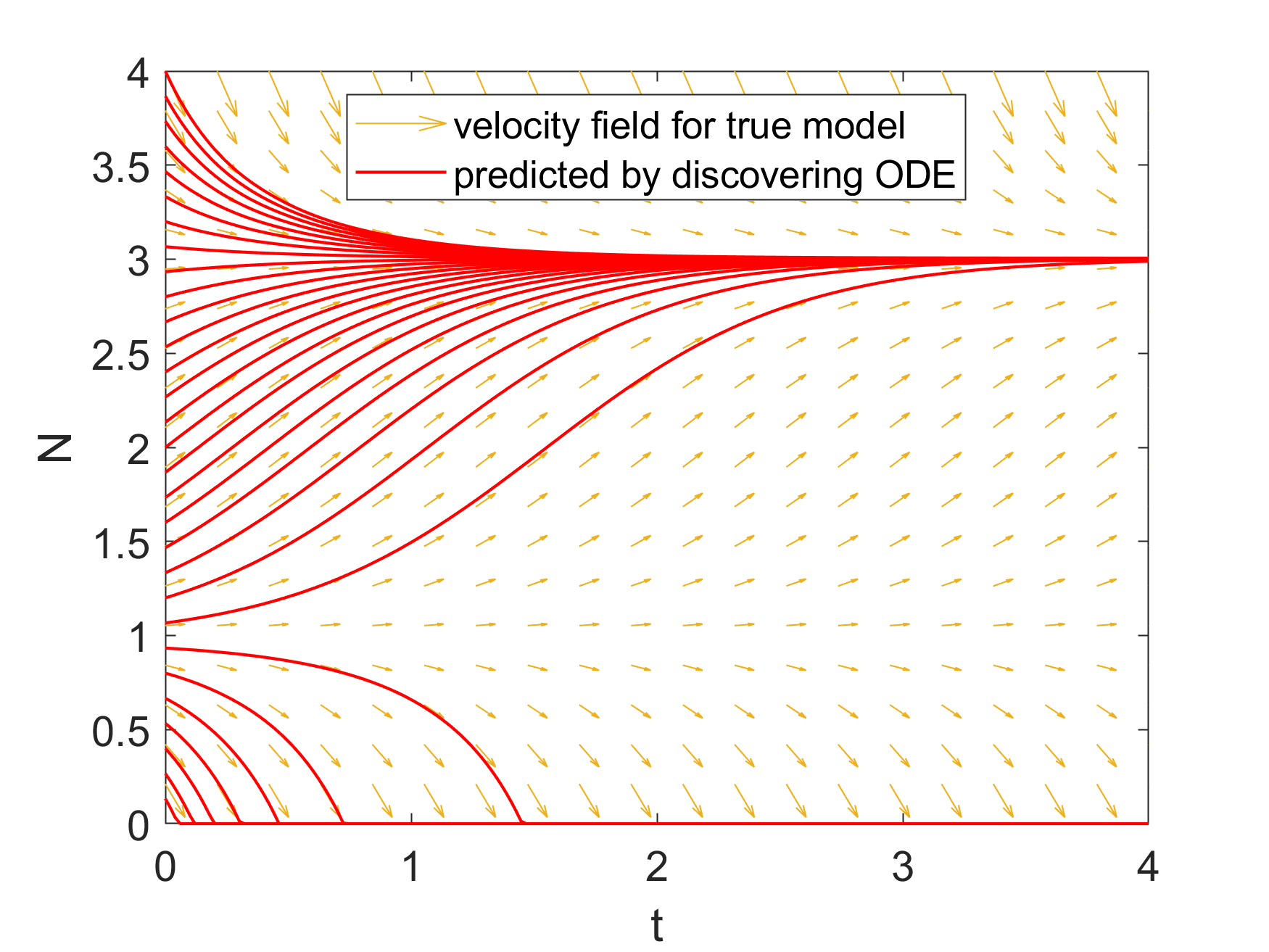}\label{fish2c}}
  (d)\subfloat{\includegraphics[width=7cm]{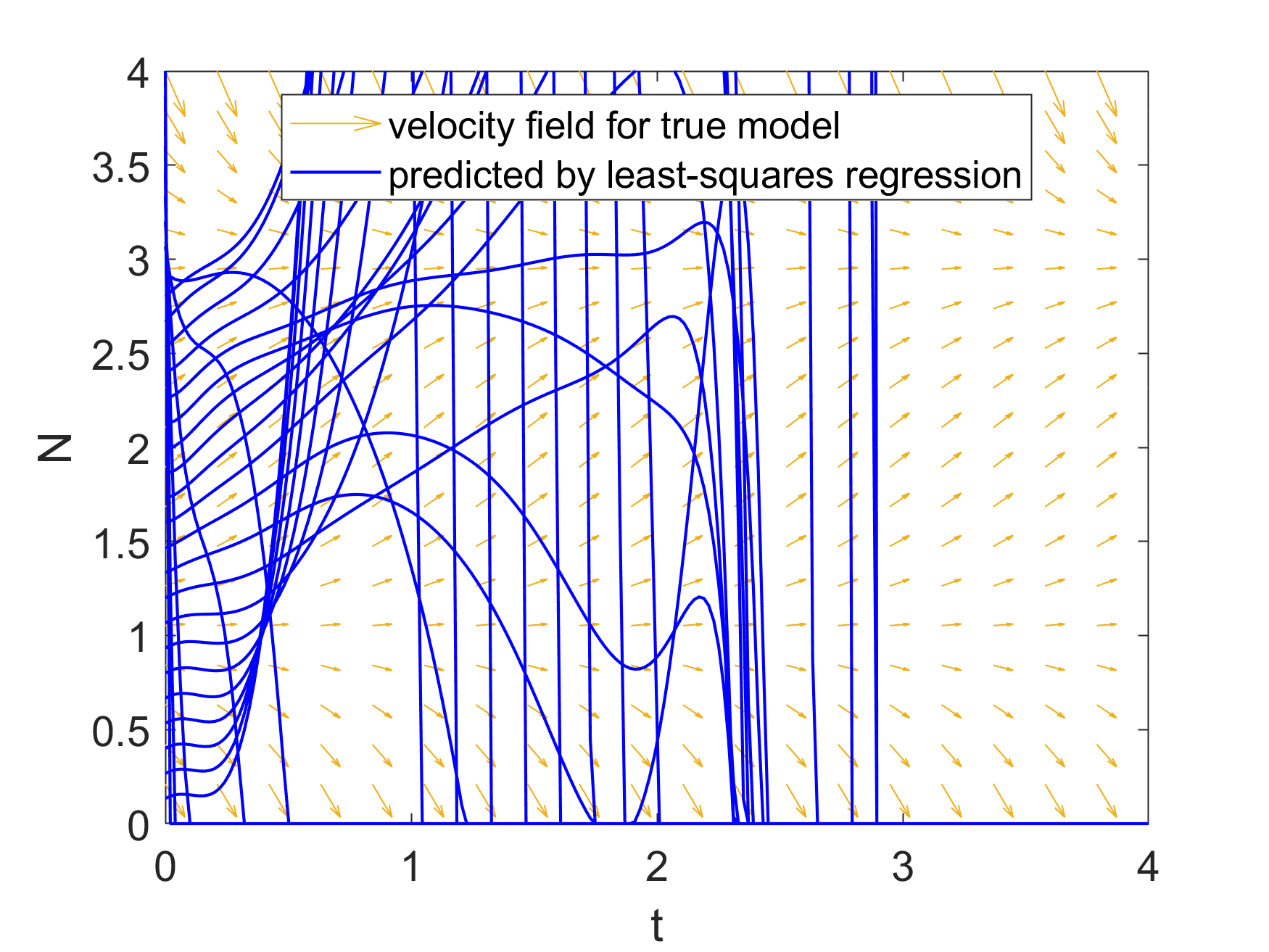}\label{fish2d}}
  \caption{[Fish-harvesting] (a) The data are generated by five random initial values and corrupted by noise. (b) The solutions to the true model (\ref{fish1}). (c) The solutions to the discovered model (\ref{fish3}). (d) The solutions calculated by least-squares regression.}
  \label{fish2}
\end{figure}

Consider the following fish-harvesting problem:
\begin{equation}
    \frac{dN}{dt} = N(4-N) - H,
\end{equation}
where $N(t)$ is the population of the fish at time $t$ and $H\ge 0$ is the constant rate at which the fish are harvested. In this example, we fix $H=3$:
\begin{equation}\label{fish1}
    \frac{dN}{dt} = N(4-N) - 3.
\end{equation}
Setting $dN/dt = 0$, we have:
\begin{equation}
    N(4-N) - 3 = 0,
\end{equation}
whose solutions are
\begin{equation}
    N = 1 \text{ and } N = 3.
\end{equation}
When the fish population $N$ is between $1$ and $3$, the population grows up; otherwise, it goes down. See Figure \ref{fish2b}.

\subsubsection{Data collection and discovery of the model}
We generate five curves on $t\in [0,2]$ using five random initial values $N_0 \sim \mathcal{U}(1,3)$ (the uniform distribution on $[1,3]$) and we collect $20$ evenly spaced data points on each curve. Then the derivatives are calculated by the five-point central-difference formula. Note that the derivatives are not calculated at the first two and last two data points on each curve. There are $80$ data points with derivative. Then all the data are corrupted by independent and identically distributed random noise $\sim \mathcal{N}(0,0.02^2)$. See Figure \ref{fish2a}. Next, we discover the ODE using our algorithm SubTSBR with subsampling size $20$, one fourth of all the data points with derivative, and $500$ subsamples. The basis-functions are monomials generated by $\{1, t, N\}$ up to degree $10$. There are $66$ terms. The result is:
\begin{equation}\label{fish3}
    \frac{dN}{dt} = -2.818 + 3.837 N - 0.965 N^2.
\end{equation}

\subsubsection{Prediction}
Now we predict the solutions using the discovered ODE (\ref{fish3}) and $30$ evenly spaced initial values $N_0$. See Figure \ref{fish2c}. Figure \ref{fish2b} shows the solutions calculated by the true model with the same initial values. As a comparison, the solutions predicted by least-squares regression are displayed in Figure \ref{fish2d}, where the data are all $100$ points illustrated in Figure \ref{fish2a}, and $u(t,N_0)$ is fitted by a linear combination of monomials generated by $\{1, t, N_0\}$ up to degree $11$ with the constraint $u(0,N_0) = N_0$. There are $66$ coefficients to estimate. Then the solutions are drawn by fixing $N_0$ at each value.

Although the data are generated by initial values $N_0$ between $1$ and $3$, the discovered model (\ref{fish3}) can be applied to other initial values and almost perfectly predicts the behavior of the true model. By contrast, least-squares regression is not able to predict well.

\section{Conclusion}
\label{summary}
In this paper, we have proposed a novel algorithm subsampling-based threshold sparse Bayesian regression (SubTSBR) to tackle high noise and outliers for data-driven discovery of differential equations. The subsampling technique has two parameters: subsampling size and the number of subsamples. When the subsampling size increases with fixed total sample size, the accuracy of SubTSBR goes up and then down. The optimal subsampling size can be fitted by the adjusted model-selection criterion. When the number of subsamples increases, the accuracy of SubTSBR keeps going up. The minimum number of subsamples needed to exclude outliers for a certain confidence level has been deduced. In practice, one may increase the number of subsamples adaptively, try different subsampling sizes, and use the adjusted model-selection criterion to choose the best subsample. SubTSBR can be used as a substitute for threshold sparse Bayesian regression (TSBR) and more accurate results may be expected. The success of SubTSBR also validates the definition of the model-selection criterion and the adjusted model-selection criterion. On top of that, we have discussed the merits of discovering differential equations from data and have demonstrated how to discover models with random initial and boundary condition as well as models with bifurcations.

\section*{Acknowledgments}
We gratefully acknowledge the support from the National Science Foundation (DMS-1555072, DMS-1736364, CMMI-1634832, and CMMI-1560834), Brookhaven National Laboratory Subcontract 382247, ARO/MURI grant W911NF-15-1-0562, and Department of Energy DE-SC0021142.

\FloatBarrier
\newpage
\bibliographystyle{elsarticle-num}
\bibliography{sections/Learn_PDE}
\end{document}